\documentclass{article}

    \PassOptionsToPackage{numbers, sort&compress}{natbib}

     \usepackage[final]{neurips_2020}

\usepackage[utf8]{inputenc} 
\usepackage[T1]{fontenc}    
\usepackage{hyperref}
\hypersetup{colorlinks=true,urlcolor=blue}      
\usepackage{url}            
\usepackage{booktabs}       
\usepackage{amsfonts}       
\usepackage{nicefrac}       
\usepackage{microtype}      
\usepackage{caption}
\newcommand{\ImageWidth}{12cm}
\usepackage{tikz}
\usetikzlibrary{decorations.pathreplacing,positioning, arrows.meta}
\usepackage{amsmath,graphicx,wrapfig} 
\usepackage{chngcntr}
\usepackage{multirow}
\newcommand{\floor}[1]{\left\lfloor #1 \right\rfloor}
\newcommand{\ceil}[1]{\left\lceil #1 \right\rceil}

\title{The Generalization-Stability Tradeoff In Neural Network Pruning}

\author{%
  Brian R. Bartoldson\thanks{Corresponding author. Majority of work completed as a student at Florida State University.} \\
  Lawrence Livermore\\
  National Laboratory\\
  \texttt{bartoldson@llnl.gov}\\
  \And
   Ari S. Morcos \\ 
 Facebook AI Research  \\ 
  \texttt{arimorcos@fb.com}
   \AND
 Adrian Barbu \\ 
 Florida State University \\
  \texttt{abarbu@stat.fsu.edu}
   \And
 Gordon Erlebacher \\
 Florida State University \\
 \texttt{gerlebacher@fsu.edu}
}

\begin{document}

\maketitle

\begin{abstract}
Pruning neural network parameters is often viewed as a means to compress models, but pruning has also been motivated by the desire to prevent overfitting. This motivation is particularly relevant given the perhaps surprising observation that a wide variety of pruning approaches increase test accuracy despite sometimes massive reductions in parameter counts. To better understand this phenomenon, we analyze the behavior of pruning over the course of training, finding that pruning's benefit to generalization \textit{increases} with pruning's instability (defined as the drop in test accuracy immediately following pruning). We demonstrate that this ``generalization-stability tradeoff'' is present across a wide variety of pruning settings and propose a mechanism for its cause: pruning regularizes similarly to noise injection. Supporting this, we find less pruning stability leads to more model flatness and the benefits of pruning do not depend on permanent parameter removal. These results explain the compatibility of pruning-based generalization improvements and the high generalization recently observed in overparameterized networks.
\end{abstract}

\section{Introduction}
\label{sec:intro}

Studies of generalization in deep neural networks (DNNs) have increasingly focused on the observation that \textit{adding} parameters improves generalization (as measured by model accuracy on previously unobserved inputs), even when the DNN already has enough parameters to fit large datasets of randomized data \citep{neyshabur2014search, zhang2016understanding}. This surprising phenomenon has been addressed by an array of empirical and theoretical analyses \citep{keskar2016large,arora2018stronger, morcos2018importance, neyshabur2018towards, belkin2018reconciling, nagarajan2019generalization,nagarajan2019deterministic,nakkiran2019deep, jiang2019fantastic,thomas2019interplay,allen2019learning}, all of which study generalization measures other than parameter counts.

Reducing memory-footprint and inference-FLOPs requirements of such well-generalizing but overparameterized DNNs is necessary to make them broadly applicable \cite{han2015deep}, and it is achievable through neural network pruning, which can substantially shrink parameter counts without harming accuracy \citep{lecun1990optimal,hassibi1993second,han2015learning,li2016pruning,he2017channel,louizos2017bayesian,liu2018rethinking}. Moreover, many pruning methods actually \textit{improve} generalization  \citep{lecun1990optimal,hassibi1993second,han2015learning,wen2016learning,narang2017exploring,liu2017learning,louizos2017learning,molchanov2017variational,frankle2018lottery,dai2018compressing,ye2018rethinking,you2019gate}. 

At the interface of pruning and generalization research, then, there's an apparent contradiction. If larger parameter counts don't increase overfitting in overparameterized DNNs, why would pruning DNN parameters throughout training improve generalization? 

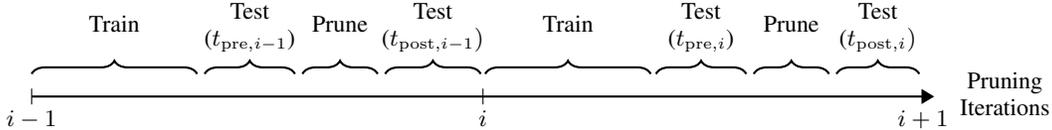
\begin{figure}[!t]
\centering
\begin{tikzpicture}
\draw[thick, -Triangle] (0,0) -- (\ImageWidth,0) node[font=\footnotesize,right=0pt]{\begin{tabular}{c} Pruning \\ Iterations \end{tabular}};

\foreach \x in {0,6}
\draw (\x cm,3pt) -- (\x cm,-3pt);

\foreach \x/\descr in {0/i-1,6/i,11.85/i+1}
\node[font=\footnotesize, text height=1.75ex,
text depth=.5ex] at (\x,-.3) {$\descr$};

\draw [thick ,decorate,decoration={brace,amplitude=5pt}] (0,0.3)  -- +(2.2,0) 
       node [black,midway,above=12pt, font=\footnotesize] {Train};
\draw [thick,decorate,decoration={brace,amplitude=5pt}] (2.3,0.3) -- +(1.2,0)
       node [black,midway,font=\footnotesize, above=4pt] {\begin{tabular}{c} Test \\ ($t_{\mathrm{pre},i-1} $) \end{tabular}};
\draw [thick,decorate,decoration={brace,amplitude=5pt}] (3.6,0.3) -- +(1,0)
       node [black,midway,font=\footnotesize, above=12pt] {Prune};
\draw [thick,decorate,decoration={brace,amplitude=5pt}] (4.7,0.3) -- +(1.27,0)
       node [black,midway,font=\footnotesize, above=4pt] {\begin{tabular}{c} Test \\ ($t_{\mathrm{post},i-1}) $ \end{tabular}};
\draw [thick ,decorate,decoration={brace,amplitude=5pt}] (6.03,0.3)  -- +(2.2,0) 
       node [black,midway,above=12pt, font=\footnotesize] {Train};
\draw [thick,decorate,decoration={brace,amplitude=5pt}] (8.3,0.3) -- +(1.2,0)
       node [black,midway,font=\footnotesize, above=4pt] {\begin{tabular}{c} Test \\ ($t_{\mathrm{pre},i} $) \end{tabular}};
\draw [thick,decorate,decoration={brace,amplitude=5pt}] (9.6,0.3) -- +(1,0)
       node [black,midway,font=\footnotesize, above=12pt] {Prune};
\draw [thick,decorate,decoration={brace,amplitude=5pt}] (10.7,0.3) -- +(1.1,0)
       node [black,midway,font=\footnotesize, above=4pt] {\begin{tabular}{c} Test \\ ($t_{\mathrm{post},i}) $ \end{tabular}};

\end{tikzpicture}
\caption{A pruning algorithm's \textit{instability} on pruning iteration $i$ is
$\mathrm{instability}_i = \frac{t_{\mathrm{pre},i} - t_{\mathrm{post},i}}{t_{\mathrm{pre},i}} $, where $t_{\mathrm{pre},i}$ and $t_{\mathrm{post},i}$ are the pruned DNN's test accuracies measured immediately before and immediately after (respectively) pruning iteration $i$. Pruning algorithm \textit{stability} on iteration $i$ is $\mathrm{stability}_i =  1- \mathrm{instability}_i$, the fraction of accuracy remaining immediately after a pruning event.}
\label{fig:timeline}
\end{figure}

We provide an answer to this question by illuminating a regularization mechanism in pruning separate from its effect on parameter counts. Specifically, we show that simple magnitude pruning \cite{han2015learning,li2016pruning} produces an effect similar to noise-injection regularization \cite{hinton1993keeping,jiang2009study,hinton2012improving,wan2013regularization,srivastava2014dropout,poole2014analyzing,neelakantan2015adding}. We explore this view of pruning as noise injection through a proxy for the level of representation ``noise'' or corruption pruning injects: the drop in accuracy immediately after a pruning event, which we call the \textit{pruning instability} (Figure \ref{fig:timeline} illustrates the computation of instability). While \textit{stability} ($\mathrm{stability} =  1- \mathrm{instability}$) is often the goal of neural network pruning because it preserves the function computed \citep{lecun1990optimal}, stable pruning could be suboptimal to the extent that pruning regularizes by noising representations during learning.

Supporting the framing of pruning as noise-injection, \textit{we find that   pruning stability is negatively correlated with the final level of generalization attained by the pruned model}. Further, this generalization-stability tradeoff appears when making changes to any of several pruning algorithm hyperparameters. For example, pruning algorithms typically prune the smallest magnitude weights to minimize their impact on network activation patterns (i.e., maximize stability). However, we observe that while pruning the \textit{largest} magnitude weights does indeed cause greater harm to stability, it also \textit{increases} generalization performance. In addition to suggesting a way to understand the repercussions of pruning algorithm design and hyperparameter choices, then, these results reinforce the idea that pruning's positive effect on DNN generalization is more about stability than final parameter count.

While the generalization-stability tradeoff suggests that pruning's generalization benefits may be present even without the permanent parameter count reduction associated with pruning, a more traditional interpretation suggests that permanent removal of parameters is critical to how pruning improves generalization. To test this, we allow pruned connections back into the network after it has adapted to pruning, and we find that the generalization benefit of permanent pruning is still obtained. This independence of pruning-based generalization improvements from permanent parameter count reduction resolves the aforementioned contradiction between pruning and generalization.

We hypothesize that lowering pruning stability (and thus adding more representation noise) helps  generalization by encouraging more flatness in the final DNN. Our experiments support this hypothesis. We find that pruning stability is negatively correlated with multiple measures of flatness that are associated with better generalization. Thus, pruning and overparameterizing may improve DNN generalization for the same reason, as flatness is also a suspected source of the unintuitively high generalization levels in overparameterized DNNs \citep{keskar2016large,neyshabur2017exploring,dziugaite2017computing,yao2018hessian,arora2018stronger,nagarajan2019deterministic,thomas2019interplay,jiang2019fantastic}.

\section{Approach}
\label{sec:approach}

Our primary aim in this work is to better understand the relationship between pruning and generalization performance, rather than the development of a new pruning method. We study this topic by varying the hyperparameters of magnitude pruning algorithms \cite{han2015learning,li2016pruning} to generate a broad array of generalization improvements and stability levels.\footnote{Our code is available at \url{https://github.com/bbartoldson/GeneralizationStabilityTradeoff}.} The generalization levels reported also reflect the generalization gap (train minus test accuracy) behavior because all training accuracies at the time of evaluation are 100\% (Section \ref{sec:iter} has exceptions that we address by plotting generalization gaps).  

In each experiment, every hyperparameter configuration was run ten times, and plots display all ten runs or a mean with 95\% confidence intervals estimated from bootstrapping.  Here, we discuss our hyperparameter choices and methodological approach. Please see Appendix \ref{sec:app_cfg} for more details.

\paragraph{Models, data, and optimization} 

We use VGG11 \citep{simonyan2014very} with batch normalization  and its dense layers replaced by a single dense layer, ResNet18, ResNet20, and ResNet56 \citep{he2016deep}. Except where noted in Section \ref{sec:iter}, we train models with Adam \citep{kingma2014adam}, which was more helpful than SGD for recovering accuracy after pruning (perhaps related to the observation  that recovery from pruning is harder when learning rates are low \cite{zhu2017prune}). We use CIFAR10 data \cite{cifar_data} without data augmentation, except in Section \ref{sec:iter} where we note use of data augmentation (random crops and horizontal flips) and Appendix \ref{sec:app_c100} where we use CIFAR100 with data augmentation to mimic  the setup in \cite{nakkiran2019deep}.  We set batch size to 128.

\paragraph{Use of $\ell_1$- and $\ell_2$-norm regularization} 

Pruning algorithms often add additional regularization via a sparsifying penalty \cite{weigend1991generalization,wen2016learning,liu2017learning,molchanov2017variational,louizos2017learning,dai2018compressing, you2019gate}, which obfuscates the intrinsic effect of pruning on generalization. Even with a simple magnitude pruning algorithm, the choice between $\ell_1$- and $\ell_2$-norm regularization affects the size of the generalization benefit of pruning \cite{han2015learning}, making it difficult to determine whether changes in generalization performance are due to changes in the pruning approach or the regularization. To avoid this confound, we study variants of simple magnitude pruning in unpenalized models, except when we note our use of the training setup of \cite{he2016deep} in Section \ref{sec:iter}.

Eschewing such regularizers may have another benefit: in a less regularized model, the size of the generalization improvement caused by pruning may be amplified. Larger effect sizes are desirable, as they help facilitate the identification of pruning algorithm facets that improve generalization. To this end, we also restrict pruning to the removal of an intermediate number of weights, which prevents pruning from harming accuracy, even when removing random or large weights \cite{li2016pruning}. 

\paragraph{Pruning schedule and rates}

For each layer of a model, the \textit{pruning schedule }specifies epochs on which pruning iterations occur (for example, two configurations in Figure \ref{fig:tradeoff_l2} prune the last VGG11 convolutional layer every 40 epochs between epochs 7 and 247). On a pruning iteration, the amount of the layer pruned is the layer's \textit{iterative pruning rate} (given as a fraction of the layer's original size), and a layer's \textit{total pruning percentage} is its iterative pruning rate multiplied by the number of scheduled pruning iterations. With the aforementioned schedule, there are seven pruning events, and a layer with total pruning percentage 90\% would have an iterative pruning rate of $\frac{90}{7}\% \approx 13\%$. Except where we note otherwise, our VGG11 and ResNet18 experiments prune just the last four convolutional layers with total pruning percentages \{30\%, 30\%, 30\%, 90\%\} and  \{25\%, 40\%, 25\%, 95\%\}, respectively. This leads to parameter reductions of 42\% for VGG11 and 46\% for ResNet18. 
 
Our experiments and earlier work \cite{bartoldson2018enhancing} indicated that focusing pruning on later layers was sufficient to create generalization and stability differences while also facilitating recovery from various kinds of pruning instability (lower total pruning percentages in earlier layers also helped recovery in \cite{li2016pruning,you2019gate}). As iterative pruning rate and schedule vary by layer to accommodate differing total pruning percentages, we note the largest iterative pruning rate used by a configuration in the plot legend. In Section \ref{sec:iter}, we test the dependence of our results on having layer-specific hyperparameter settings by pruning 10\% of every layer in every block of ResNet18, ResNet20, and ResNet56. 

\paragraph{Parameter scoring and pruning target}
We remove entire filters (structured pruning), and we typically \textit{score} filters of VGG11 using their $\ell_2$-norm and filters of ResNet18---which has feature map shortcuts not accounted for by filters---using their resulting feature map activations' $\ell_1$-norms \cite{polyak2015channel,li2016pruning}, which we compute with a moving average. Experiments in Section \ref{sec:iter}, Appendix \ref{sec:app_trade}, and Appendix \ref{sec:app_c100} use other scoring approaches, including  $\ell_1$-norm scoring of ResNet filters in Section \ref{sec:iter}. We denote pruning algorithms that \textit{target}/remove the smallest-magnitude (lowest-scored) parameters with an "S" subscript (e.g. Prune$_\mathrm{S}$ or Prune\_S), random parameters with an "R" subscript, and the largest-magnitude parameters with an "L" subscript. Please see Appendix \ref{sec:app_cfg} for more pruning details.

\paragraph{Framing pruning as noise injection}  Pruning is typically a deterministic procedure, with the weights that are targeted for pruning being defined by a criterion (e.g., the bottom 1\% of weights in magnitude). Given weights meeting such a criterion, pruning can be effected through their multiplication by a $\mathrm{Bernoulli}(p)$ distributed random variable, where $p=0$. Setting $p > 0$ would correspond to DropConnect, a DNN noise injection approach and generalization of dropout \cite{hinton2012improving,wan2013regularization,srivastava2014dropout}. Thus, for weights meeting the pruning criterion, pruning is a limiting case of a noise injection technique. Since not all weights matter equally to a DNN's computations, we measure the amount/salience of the ``noise'' injected by pruning via the drop in accuracy immediately following pruning (see Figure \ref{fig:timeline}).

In Section \ref{sec:exp_noise}, we show that pruning's generalization benefit can be obtained without permanently removing parameters. Primarily, we achieve this by multiplying by zero---for a denoted number of training batches---the parameters we would normally prune, then returning them to the model (we run variants where they return initialized at the values they trained to prior to zeroing, and at zero as in \cite{han2016dsd}). In a separate experiment, we replace the multiplication by zero with the addition of Gaussian noise, which has a variance equal to the variance of the unperturbed parameters on each training batch and a larger variance on the first batch of a new epoch. Please see Appendix \ref{sec:app_noise} for more details.

\paragraph{Computing flatness} In Section \ref{sec:flatness}, we use test data \cite{thomas2019interplay} to compute approximations to the traces of the Hessian of the loss $\mathbf{H}$ (curvature) and the gradient covariance matrix $\mathbf{C}$ (noise).\footnote{We use ``flatness'' loosely when discussing the trace of the gradient covariance, which is large/``sharp'' when the model's gradient is very sensitive to changes in the data sample and small/``flat'' otherwise.} $\mathbf{H}$ indicates the gradient's sensitivity to parameter changes at a point, while $\mathbf{C}$ shows the sensitivity of the gradient to changes in the sampled input (see Figure \ref{fig:flatness}) \cite{thomas2019interplay}. The combination of these two matrices via the Takeuchi information criterion (TIC) \cite{takeuchi1976distribution} is particularly predictive of generalization \cite{thomas2019interplay}. Thus, in addition to looking at $\mathbf{H}$ and/or $\mathbf{C}$ individually, as has been done in \cite{yao2018hessian,jiang2019fantastic}, we also consider a rough TIC proxy $\mathrm{Tr}(\mathbf{C})/\mathrm{Tr}(\mathbf{H})$ inspired by \cite{thomas2019interplay}. Finally, similar to analyses in \cite{keskar2016large,yao2018hessian,jiang2019fantastic}, we compute the size $\varepsilon$ of the parameter perturbation (in the directions of the Hessian's dominant eigenvectors) that can be withstood before the loss increases by 0.1.

\section{Experiments}
\label{sec:exp}

\subsection{The generalization-stability tradeoff}
\label{sec:exp_gen}

Can improved generalization in pruned DNNs simply be explained by the reduced parameter count, or rather, do the properties of the pruning algorithm play an important role in the resultant generalization? As removing parameters from a DNN via pruning may make the DNN less capable of fitting to the noise in the training data \cite{lecun1990optimal,hassibi1993second,liu2018rethinking}, we might expect that the generalization improvements observed in pruned DNNs are entirely explained by the number of parameters removed at each layer. In which case, methods that prune equal amounts of parameters per layer would generalize similarly.

Alternatively, the nature of the particular pruning algorithm might determine generalization improvements. While all common pruning approaches seek to preserve important components of the function computed by the overparameterized DNN, they do this with varying degrees of success, creating different levels of stability. More stable approaches include those that compute a very close approximation to the way the loss changes with respect to each parameter and prune a single parameter at a time \citep{hassibi1993second}, while less stable approaches include those that assume parameter magnitude and importance are roughly similar and prune many weights all at once \citep{han2015learning}. Therefore, to the extent that differences in the noise injected by pruning explain differences in pruning-based generalization improvements, we might expect to observe a relationship between generalization and pruning stability.

\begin{figure}[t]
  \centering
    \begin{minipage}[!t]{.65\linewidth}
      \centering
     \centerline{\includegraphics[width=\linewidth]{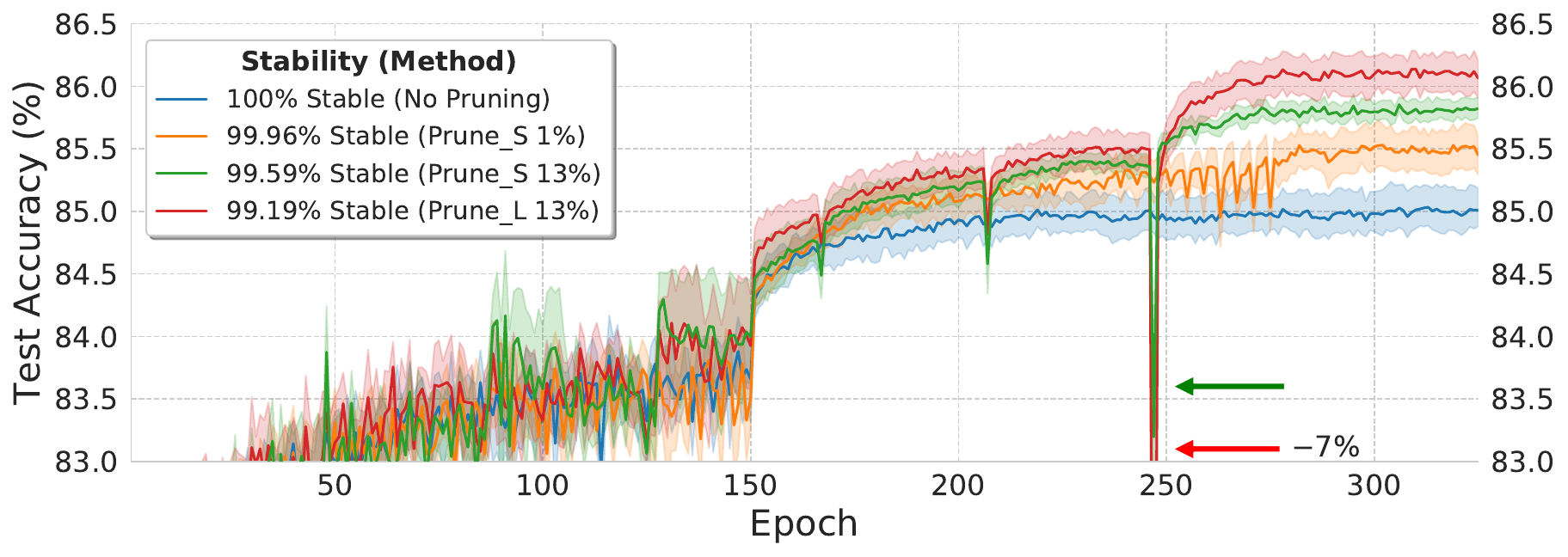}}
    \end{minipage}
    \begin{minipage}[!t]{.34\linewidth}
      \centering
        \centerline{\includegraphics[width=\linewidth]{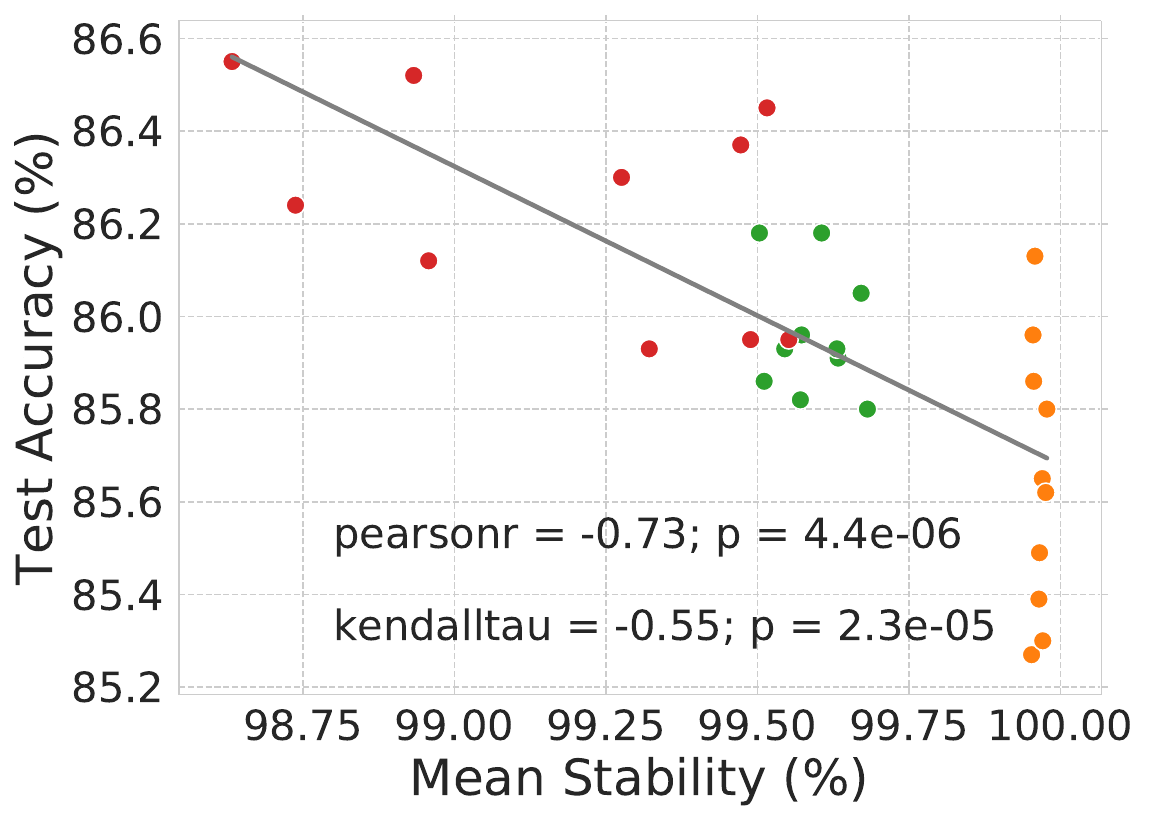}}
    \end{minipage}
    \begin{minipage}[!t]{.65\linewidth}
      \centering
     \centerline{\includegraphics[width=\linewidth]{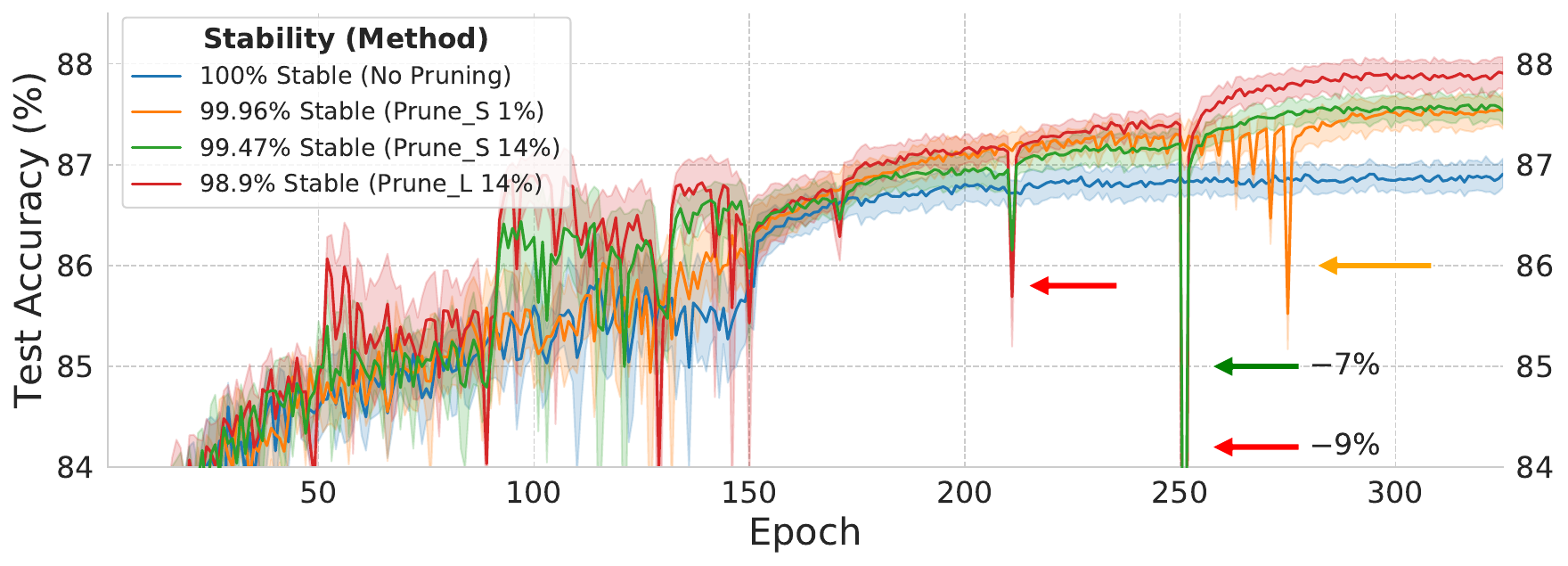}}
    \end{minipage}
    \begin{minipage}[!t]{.34\linewidth}
      \centering
        \centerline{\includegraphics[width=\linewidth]{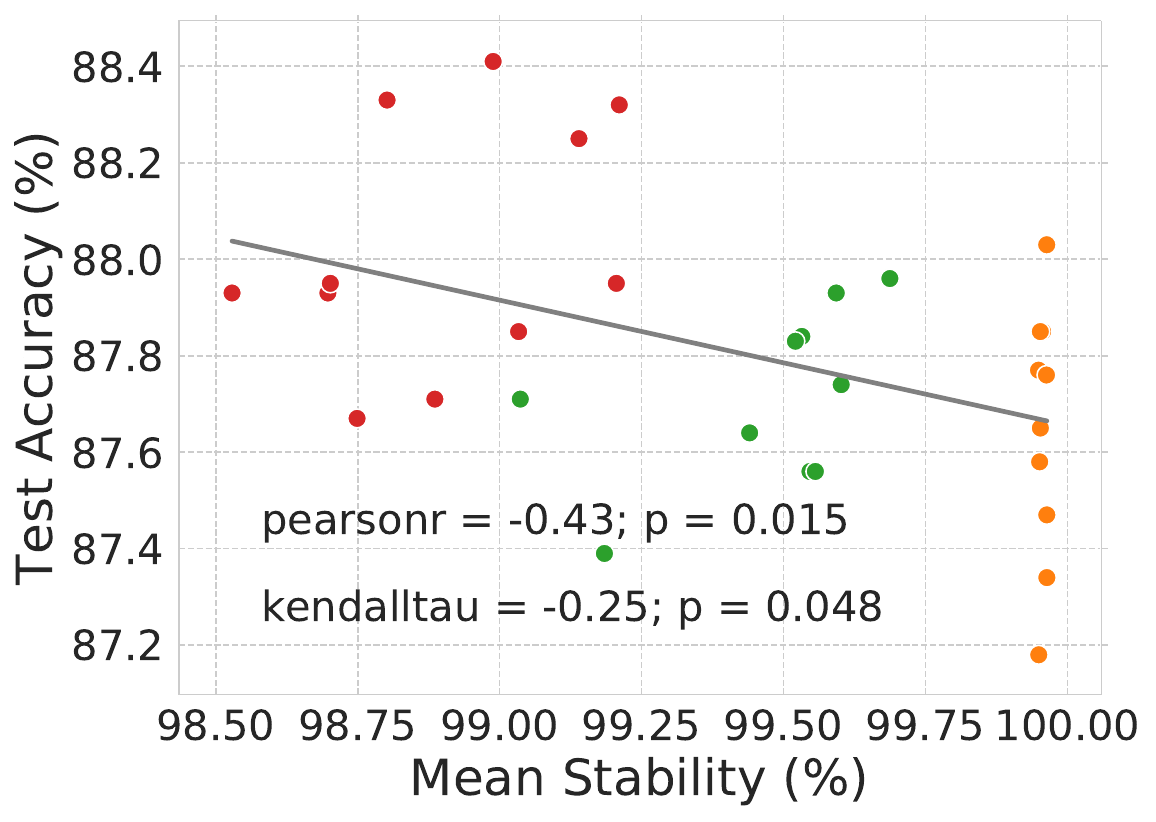}}
    \end{minipage}
\caption{Less stable pruning leads to higher generalization in VGG11 (top)  and ResNet18 (bottom)  when training on CIFAR-10 (10 runs per configuration). (Left) Test accuracy during training of several models illustrates how adaptation to less stable pruning leads to better generalization. (Right) Means reduce along the epoch dimension (creating one point per run-configuration combination).}
\label{fig:tradeoff_l2}
\end{figure}

To determine whether pruning algorithm stability affects generalization, we compared the stability and final test accuracy of several pruning algorithms with varying pruning targets and iterative pruning rates (Figure \ref{fig:tradeoff_l2}). Consistent with the nature of the pruning algorithm playing a role in generalization, we observed that \textit{less stable pruning algorithms created higher final test accuracies} than those which were stable (Figure \ref{fig:tradeoff_l2}, right; VGG11: Pearson's correlation $r=-.73$, p-value $=4.4\mathrm{e}{-6}$; ResNet18: $r=-.43$, p-value $=.015$). While many pruning approaches have aimed to be as stable as possible, these results suggest that pruning techniques may actually facilitate better generalization when they induce \emph{less} stability. In other words there is a tradeoff between the stability during training and the resultant generalization of the model. Furthermore, these results show that parameter-count- and architecture-based \cite{liu2018rethinking} arguments are not sufficient to explain generalization levels in pruned DNNs, as the precise pruning method plays a critical role in this process.

Figure \ref{fig:tradeoff_l2} also demonstrates that pruning events for Prune$_\mathrm{L}$ with a high iterative pruning rate (red curve, pruning as much as 14\% of a given convolutional layer per pruning iteration) are substantially more destabilizing than other pruning events, but despite the dramatic pruning-induced drops in performance, the network recovers to higher performance within a few epochs. Several of these pruning events are highlighted with red arrows. Please see Appendix \ref{sec:app_trade} for more details.

Appendix \ref{sec:app_trade} also shows results with a novel scoring method that led to a wider range of stabilities and generalization levels, which improved the correlations between generalization and stability in both DNNs. Thus, the visibility of the generalization-stability tradeoff is affected by pruning algorithm hyperparameter settings, accenting the benefit of designing experiments to allow large pruning-based generalization gains. In addition, these results suggest that the regularization levels associated with various pruning hyperparameter choices may be predicted by their effects on stability during training.

\subsection{Towards understanding the bounds of the generalization-stability tradeoff}
\label{sec:iter}

In Figure \ref{fig:tradeoff_l2}, decreasing pruning algorithm stability led to higher final generalization. Will decreasing stability always help generalization? Is the benefit of instability present in smaller DNNs and when training with SGD? Here, we address these and similar questions and ultimately find that the tradeoff has predictable limits but is nonetheless present across a wide range of experimental hyperparameters.

\begin{figure}[!t]
    \centering
    \begin{minipage}[!t]{.285\linewidth}
      \centering
  \centerline{\includegraphics[width=\linewidth]{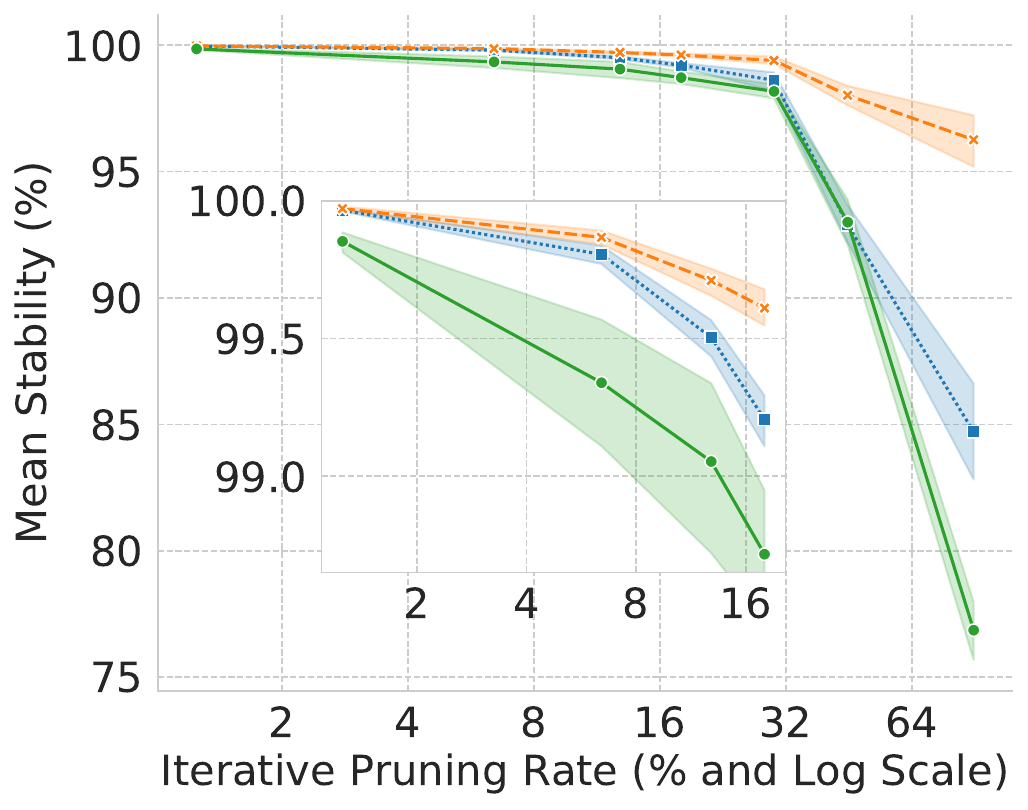}}
    \end{minipage}
    \begin{minipage}[!t]{.40\linewidth}
      \centering
  \centerline{\includegraphics[width=\linewidth]{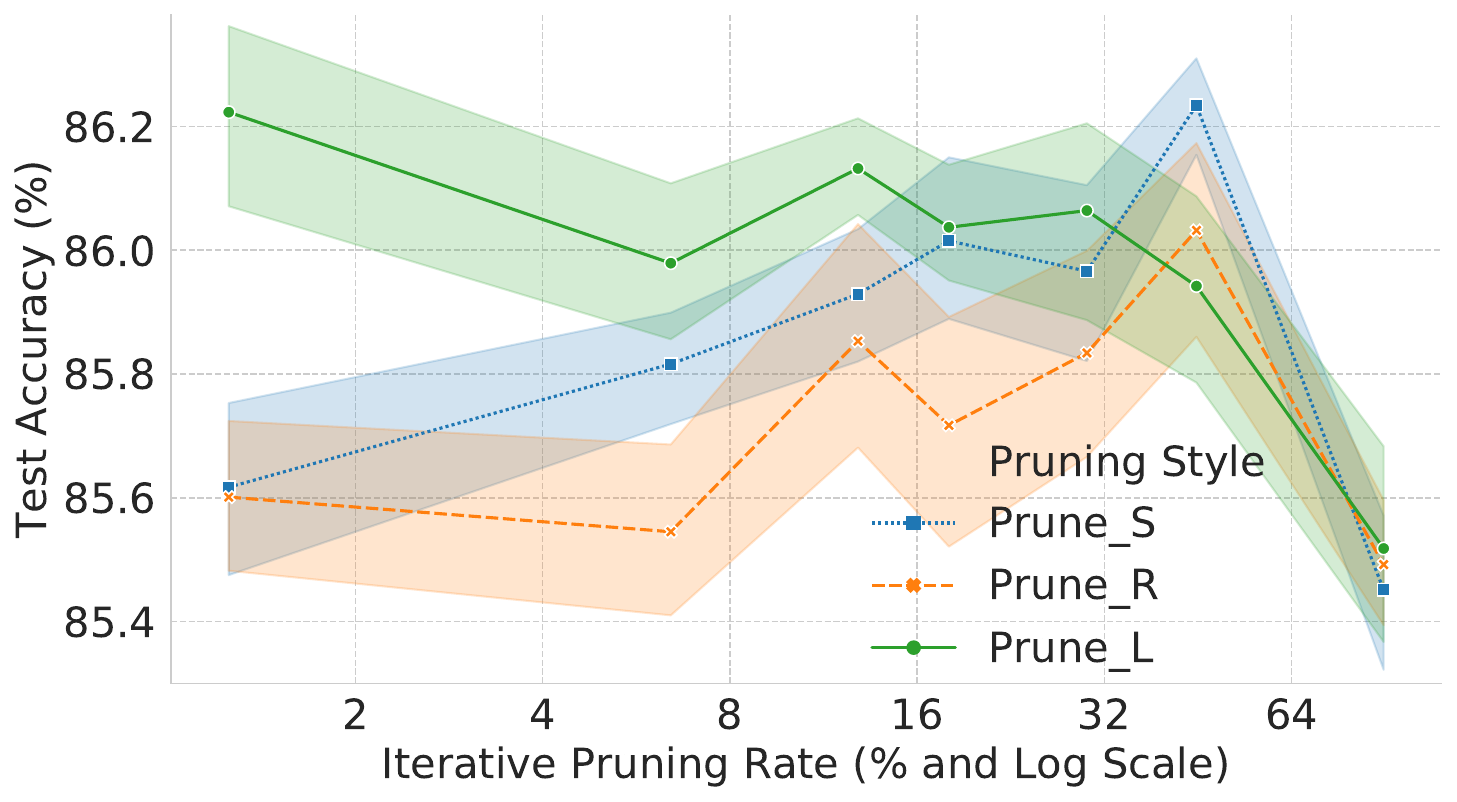}}
    \end{minipage}
    \begin{minipage}[!h]{.3\linewidth}
      \centering
  \centerline{\includegraphics[width=\linewidth]{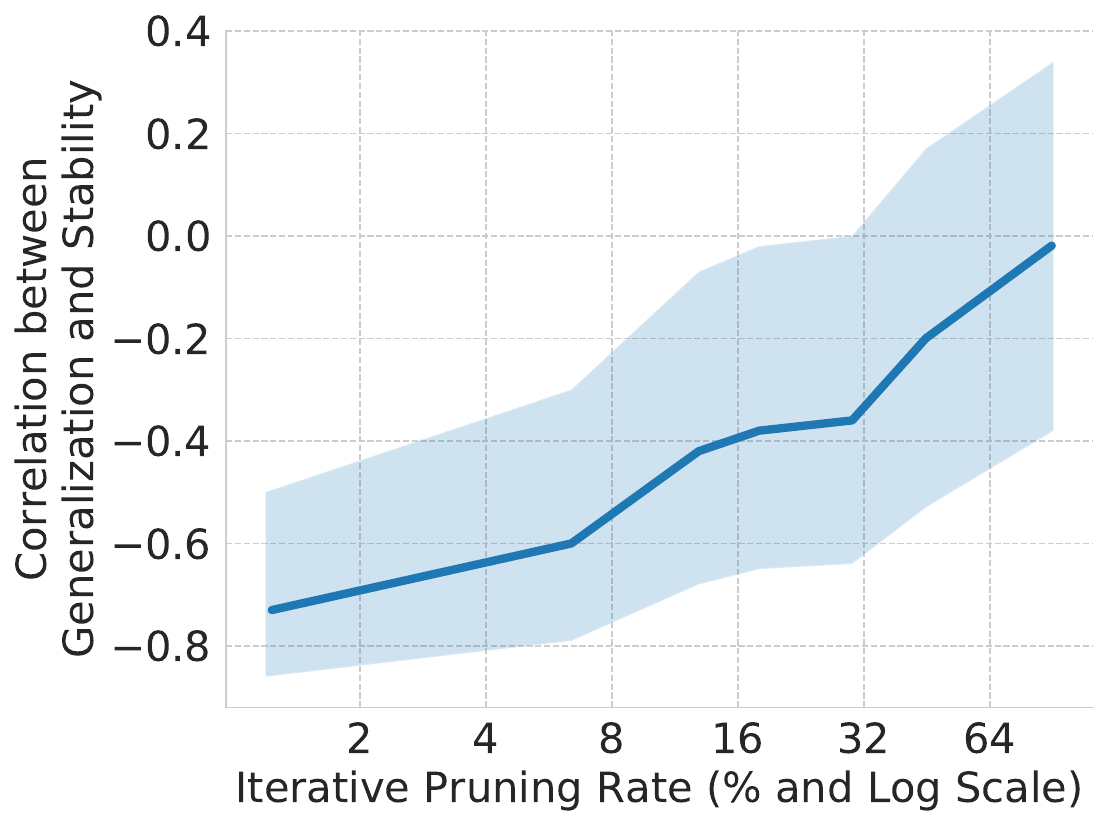}}
    \end{minipage}

\caption{Increasing the iterative pruning rate (and decreasing the number of pruning events to hold total pruning constant) leads to less stability (left), and can allow methods that target less important parameters to generalize better (center). At a particular iterative rate, the Pearson correlation between generalization and stability is always negative (right), a similar pattern holds with Kendall's rank correlation. A baseline has 85.2\% accuracy. }
\label{fig:retrain}
\end{figure}

\paragraph{Impact of iterative pruning rate on the generalization-stability tradeoff} 
For a particular pruning target and total pruning percentage, pruning stability in VGG11 monotonically decreases as we raise the iterative pruning rate up to the maximal, one-shot-pruning level (Figure \ref{fig:retrain} left). Thus, if less stability is always better, we would expect to see monotonically increasing generalization as we raise iterative pruning rate. Alternatively, it's possible that we will observe a generalization-stability tradeoff over a particular range of iterative rates, but that there will be a point at which lowering stability further will not be helpful to generalization. To test this, we compare iterative pruning rate and test accuracy for each of three pruning targets (Figure \ref{fig:retrain} center). 

For pruning targets that are initially highly stable (Prune$_\mathrm{S}$ and Prune$_\mathrm{R}$), raising the iterative pruning rate and decreasing stability produces higher generalization up until the one-shot pruning case (Figure \ref{fig:retrain} center). When the pruning target lacks stability at the initial iterative rate (Prune$_\mathrm{L}$), further decreasing stability is \textit{harmful} to generalization. These results suggest that the generalization stability tradeoff is present across a wide range of iterative pruning rates, but, critically, that there are limits to the benefits of further decreasing stability once it is already at a low level.  

Interestingly, we found that the generalization-stability tradeoff grew \textit{weaker} as the iterative pruning rate increased as well (Figure \ref{fig:retrain} right). Notably, however, the tradeoff was present for all iterative pruning rates studied (though at the highest iterative rates, the correlation is no longer significant). This result suggests that not only does the generalization improvement decrease as stability decreases past some threshold, the strength of the \textit{tradeoff itself} also decreases as stability decreases, highlighting that there is a ``sweet spot'' at which decreased stability is most helpful.

\paragraph{Impact of traditional training and pruning on the generalization-stability tradeoff}

Our experiments thus far (e.g. those shown in Figure \ref{fig:tradeoff_l2}) pruned only a subset of layers of large models trained with Adam, without weight decay or data augmentation. It's possible that reductions in stability only improve generalization in such a regime. Alternatively, the tradeoff may be present when making changes to these factors.

We investigate this important matter by evaluating the relationship between generalization and stability in ResNet18, ResNet20, and ResNet56 when training using the hyperparameters described in \cite{he2016deep} (e.g., we employ SGD with weight decay and data augmentation). Further, we simplify our pruning approach by removing 10\% of the filters of each convolutional layer of each block, scoring filters with their $\ell_1$-norms. Parameters are removed either three times during training (epochs \{41, 71, 101\}) or twice during training (epochs \{41, 101\}), creating iterative rates of roughly 3\% and 5\%.

\begin{figure}[t]
  \centering
    \begin{minipage}[!t]{.3\linewidth}
      \centering
    \scriptsize{\textbf{ResNet20}\par\smallskip}
     \centerline{\includegraphics[width=\linewidth]{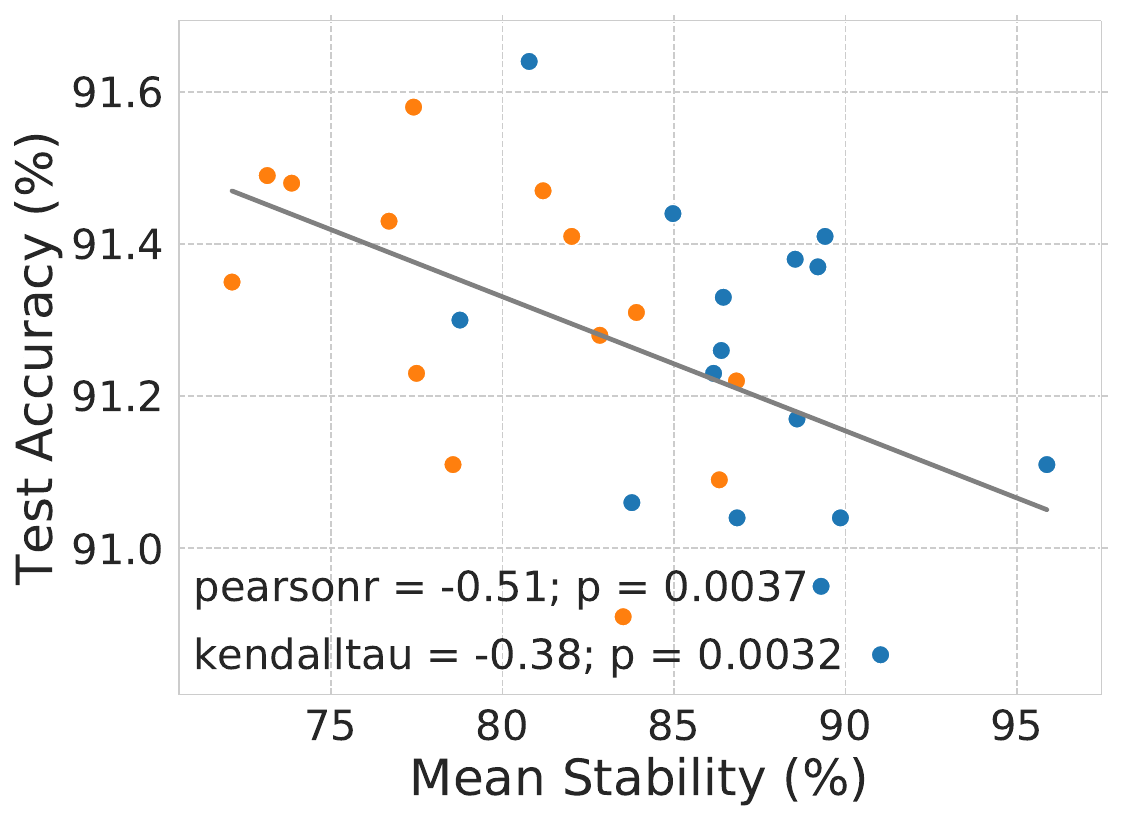}}
    \end{minipage}
    \begin{minipage}[!t]{.03\linewidth}
    \ 
    \end{minipage}
    \begin{minipage}[!t]{.3\linewidth}
      \centering
    \scriptsize{\textbf{ResNet56}\par\smallskip}
     \centerline{\includegraphics[width=\linewidth]{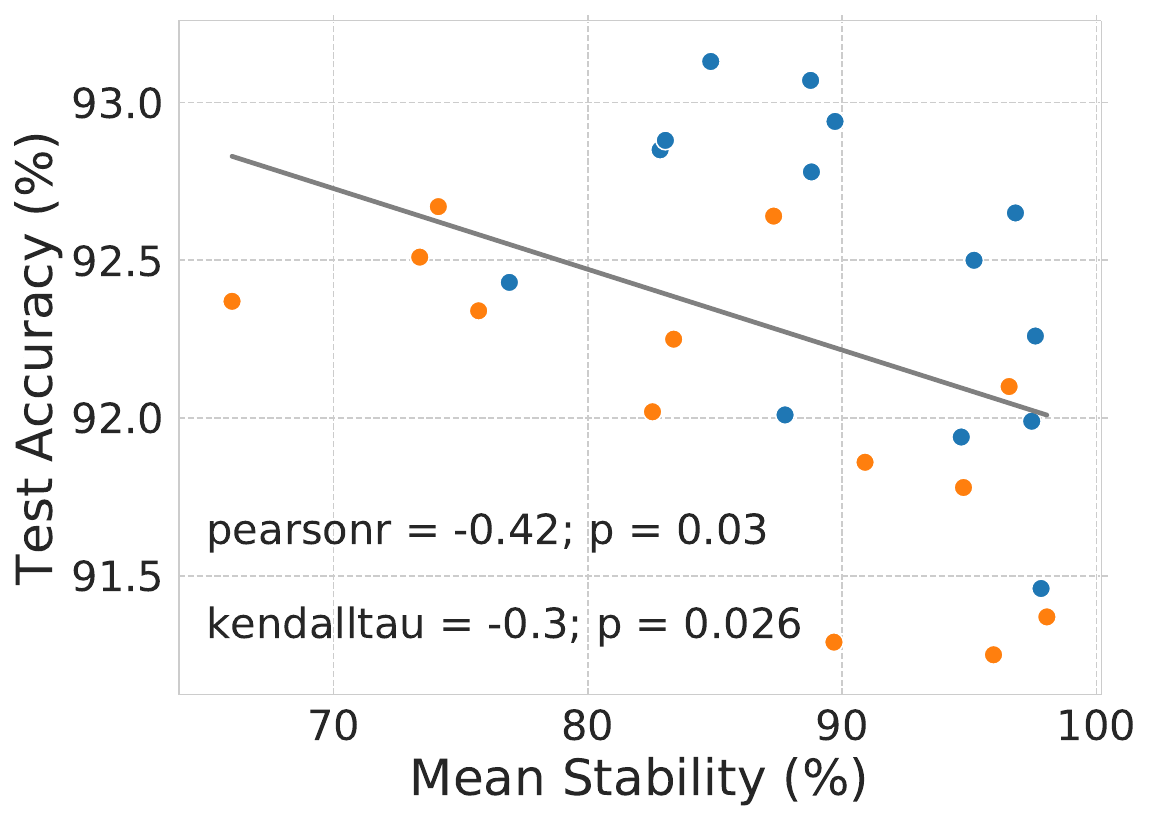}}
    \end{minipage}
    \begin{minipage}[!t]{.03\linewidth}
    \ 
    \end{minipage}
    \begin{minipage}[!t]{.3\linewidth}
      \centering
    \scriptsize{\textbf{ResNet18}\par\smallskip} 
     \centerline{\includegraphics[width=\linewidth]{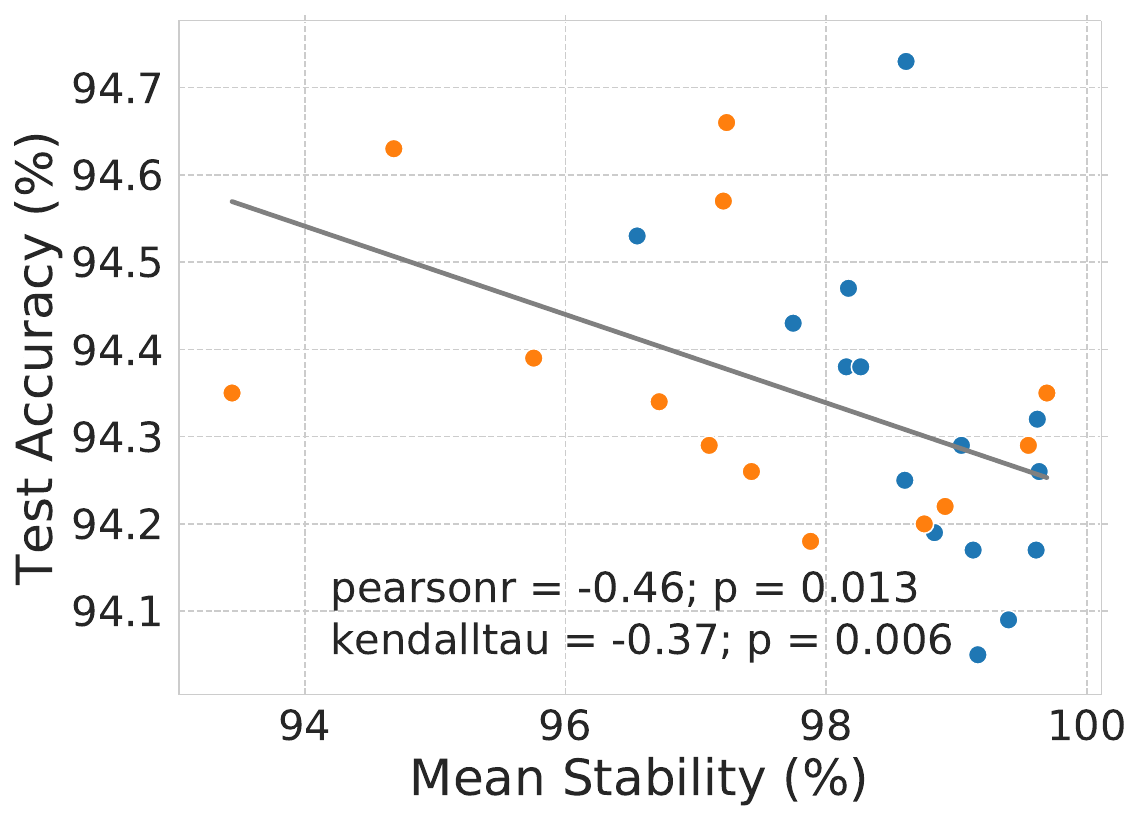}}
    \end{minipage}
    \begin{minipage}[!t]{.3\linewidth}
      \centering
     \centerline{\includegraphics[width=\linewidth]{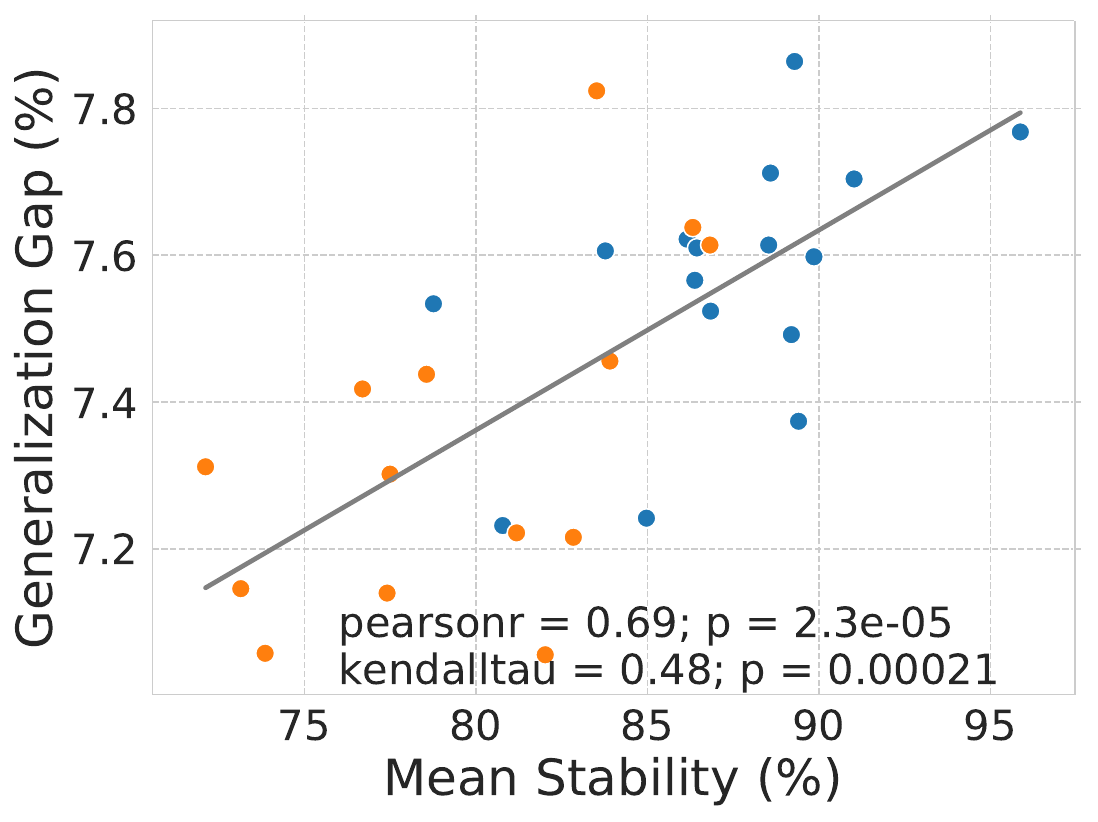}}
    \end{minipage}
    \begin{minipage}[!t]{.03\linewidth}
    \ 
    \end{minipage}
    \begin{minipage}[!t]{.3\linewidth}
      \centering
     \centerline{\includegraphics[width=\linewidth]{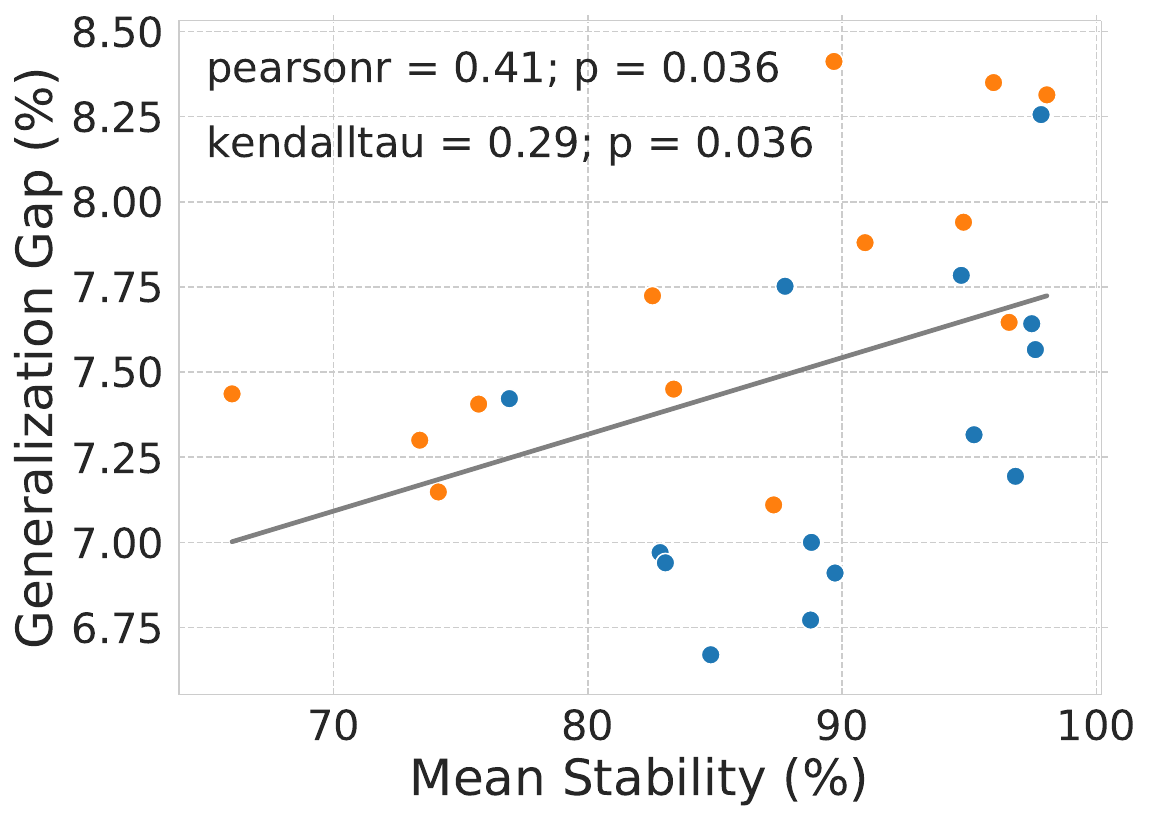}}
    \end{minipage}
    \begin{minipage}[!t]{.03\linewidth}
    \ 
    \end{minipage}
    \begin{minipage}[!t]{.3\linewidth}
      \centering
     \centerline{\includegraphics[width=\linewidth]{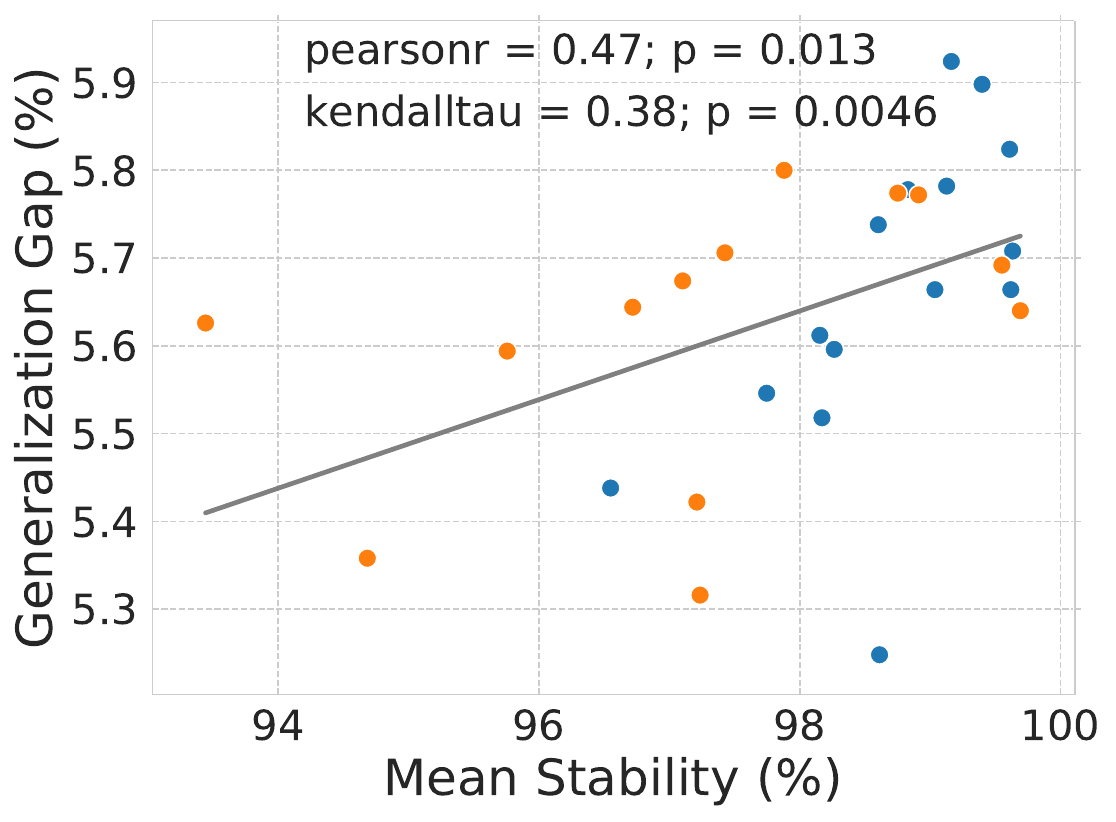}}
    \end{minipage}
\caption{Among pruned models, lower pruning stability is associated with higher generalization and lower generalization gaps (overfitting) in ResNet18, ResNet20, and ResNet56 when training with weight decay and data augmentation. Blue and orange dots represent models pruned with 3\% and 5\% iterative rates, respectively.
}
\label{fig:resnetM}
\end{figure}

Consistent with the generalization-stability tradeoff explaining generalization levels across various training and pruning scenarios, Figure \ref{fig:resnetM} shows that reductions in stability improve both generalization and the generalization gap in pruned models. In Appendix \ref{sec:app_traditional}, we build on these results and show a stability regime where lower stability leads to  generalization levels higher than the baseline model's.

\paragraph{Impact of total pruning percentage on the generalization-stability tradeoff}
We raised the total pruning percentage in the Figure \ref{fig:tradeoff_l2} ResNet18 experiments from 46\% to 59\% and found that the generalization-stability tradeoff was still present. Interestingly, however, Prune$_\mathrm{L}$ seemingly induced too much instability and ceased to outperform Prune$_\mathrm{S}$ at this higher total pruning percentage, consistent with prior work \cite{li2016pruning} which found that pruning large weights was harmful. Please see Appendix \ref{sec:app_hyper} for these additional results and more details of the experiments in this section.

Taken together, these results demonstrate that while the generalization-stability tradeoff was present across a wide range of pruning hyperparameters, it consistently broke down once pruning stability dropped below some threshold, at which point further reducing stability did not lead to generalization improvements. This failure mode highlights the need to frame the benefits of lower stability as a part of a tradeoff rather than a free lunch. Further, it is consistent with the comparison to noise-injection, wherein the noise is moderate (e.g., increasing the dropout rate past 0.8 harms generalization) \citep{hinton1993keeping,jiang2009study,hinton2012improving,wan2013regularization,srivastava2014dropout,poole2014analyzing,neelakantan2015adding}.

\subsection{Iterative magnitude pruning as noise injection}
\label{sec:exp_noise}

We have alluded to the idea that simple magnitude pruning performs a kind of noise injection, with the peculiarity that the noise is applied permanently or not at all. Removing the permanence of pruning by allowing weight reentry can mitigate the parameter reduction of pruning, making it more similar to a traditional noise-injection regularizer, and allowing us to test whether the permanent reduction in parameters caused by pruning is critical to its effect on generalization.

\begin{wrapfigure}[26]{r}{0.46\linewidth}
    \centering
    \begin{minipage}[!t]{.95\linewidth}
      \centering
  \centerline{\includegraphics[width=\linewidth]{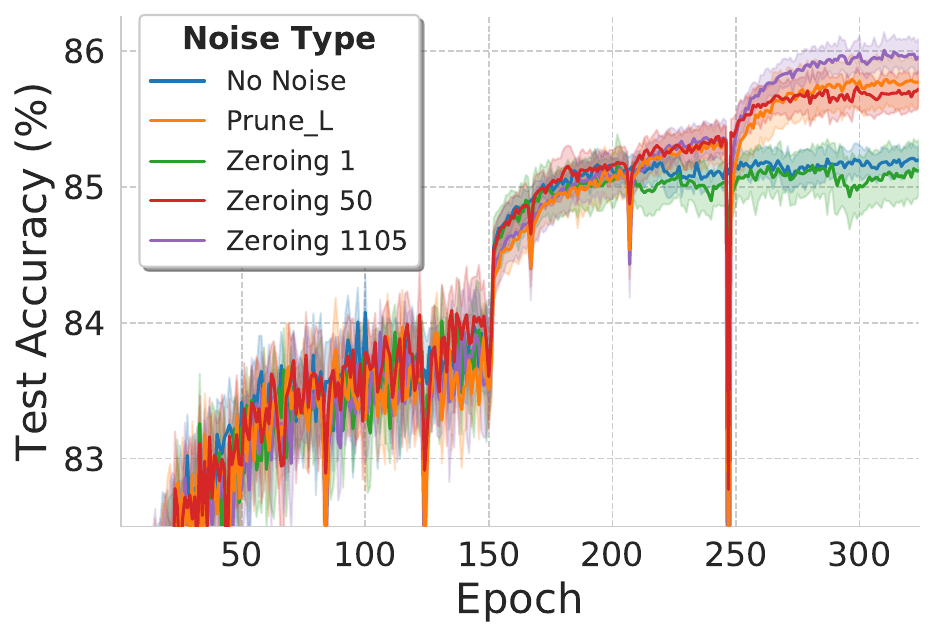}}
    \end{minipage}
    \begin{minipage}[!t]{.95\linewidth}
      \centering
  \centerline{\includegraphics[width=\linewidth]{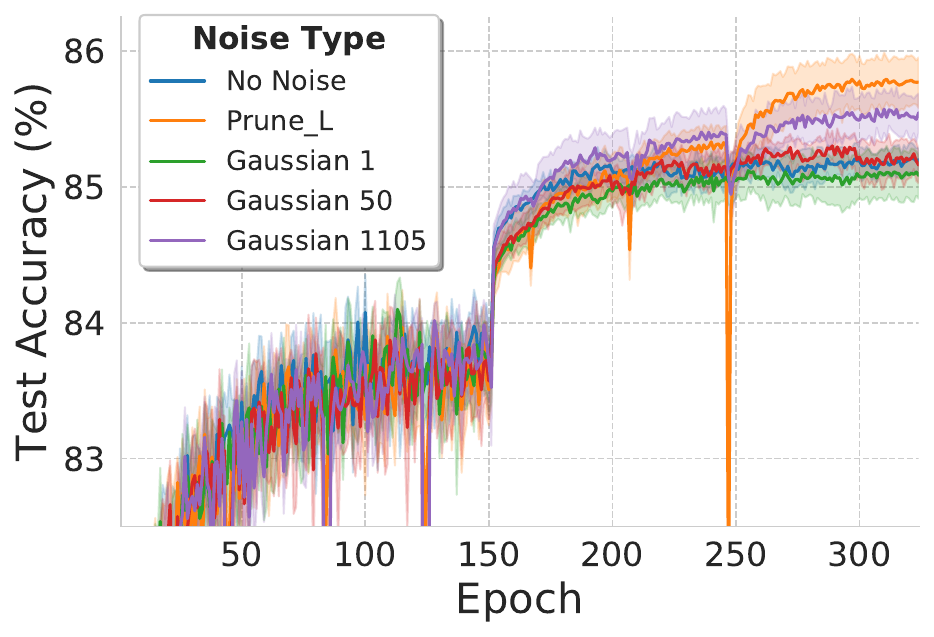}}
    \end{minipage}
    \vspace{-2mm}
\caption{Generalization improvements from pruning bear resemblance to those obtained by using temporary multiplicative zeroing (top)  and  additive Gaussian noise (bottom), as long as the noise is applied for enough batches/steps.}
\label{fig:noise}
\end{wrapfigure}

As a baseline, we consider Prune$_\mathrm{L}$ applied to VGG11, judging filter magnitude via the $\ell_2$-norm. We then modify this algorithm to, rather than permanently prune filters, simply set the filter weights to zero, then allow the zeroed weights to immediately resume training in the network ("Zeroing 1" in Figure \ref{fig:noise} top). However, by allowing pruned weights to immediately recover, Zeroing 1 differs from pruning noise, which causes the unpruned features to be trained in the absence of the pruned feature maps.

\setlength\intextsep{0pt}

To retain this potentially regularizing aspect of pruning noise, we held weights to zero for 50 and 1105 consecutive batches, as well. As a related experiment, we measured the impact of adding Gaussian noise to the weights either once (Gaussian 1) or repeatedly over a series of training batches (Gaussian 50/1105 in Figure \ref{fig:noise} bottom). 

If the capacity reduction associated with having fewer parameters is not necessary to explain pruning's effect on generalization, then we would expect that the generalization behavior of temporary pruning noise injection algorithms could mimic the generalization behavior of  Prune$_\mathrm{L}$. Alternatively, if having fewer weights is a necessary component of pruning-based generalization improvements, then we would not expect close similarities between the generalization phenomena of  Prune$_\mathrm{L}$ and temporary pruning noise injection. 

Consistent with the idea that the noise injected by pruning leads to the generalization benefits observed in pruned DNNs, applying zeroing noise for 50 batches to filters (rather than pruning them completely) generates strikingly similar final generalization performance to Prune$_\mathrm{L}$ (Figure \ref{fig:noise} top). In fact, throughout training, both methods have similar levels of instability and test accuracy. This result suggests that pruning-based generalization improvements in overparameterized DNNs do not require the model's parameter count to be reduced. 

Finally, we evaluated the impact of adding Gaussian noise to parameters at various points throughout training. Consistent with the generalization-stability tradeoff, we found that when Gaussian noise was added for a long enough duration (Gaussian 1105; purple line in Figure \ref{fig:noise} bottom), performance increased substantially. This result demonstrates that the generalization-stability tradeoff is not specific to pruning, and that noise injected by pruning is simply a special case of noise more broadly.

Additional results and experimental details are in Appendix \ref{sec:app_noise}. For example, an alternative version of this analysis zeros weights for N batches, then allows them back in at their pre-zeroing values. This method creates instability similar to regular pruning's, and produces a similar generalization benefit. Also, we provide a visualization of the weight noising methods that we use here.

\subsection{Flatness: a mechanism for pruning-based generalization improvements?}
\label{sec:flatness}

Our results thus far suggest that noise injection is the mechanism through which pruning improves generalization. Can the noise pruning adds to representations translate to flatness in the final model that improves generalization? Here, we address this question.

Given the many successful versions of noise injection \cite{hinton1993keeping,jiang2009study,hinton2012improving,wan2013regularization,srivastava2014dropout,poole2014analyzing,neelakantan2015adding}, and pruning's relationship to dropout \cite{leclerc2018smallify,gomez2018targeted}, we hypothesize that pruning noise can produce flatness in the resulting model that's helpful to generalization. Specifically, we expect that less stable pruning, which introduces more significant noise by definition, will translate to heightened model robustness to changes in data sample and parameters (flatness). Furthermore, we expect that the heightened flatness will translate to higher generalization, consistent with empirical evidence and theory suggesting that flatness is helpful to generalization in overparameterized DNNs \cite{keskar2016large,chaudhari2016entropy,dziugaite2017computing,neyshabur2017exploring,arora2018stronger,yao2018hessian,jiang2019fantastic,thomas2019interplay}.

Alternatively, it's possible that the observed relationship between pruning stability and generalization is merely correlation, that pruning noise helps in a way unrelated to flatness, or that flatness differences don't explain the generalization benefits created by pruning. If we observed a positive or no correlation between stability and flatness, or a negative correlation between flatness and generalization, then our experiments would support one of these alternative hypotheses.

\begin{figure}[t]
  \centering
    \begin{minipage}[!t]{.4\linewidth}
      \centering
     \centerline{\includegraphics[width=\linewidth]{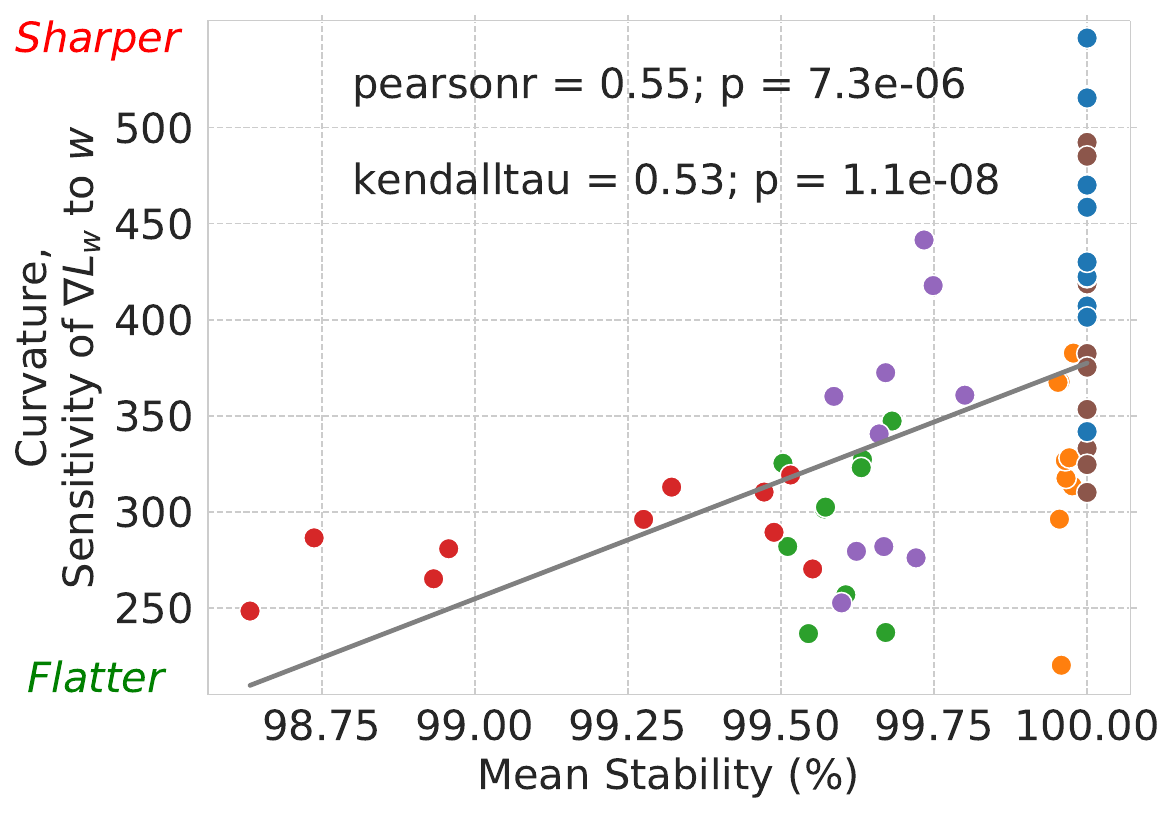}}
    \end{minipage}
    \begin{minipage}[!t]{.09\linewidth}
    \ 
    \end{minipage}
    \begin{minipage}[!t]{.4\linewidth}
      \centering
        \centerline{\includegraphics[width=\linewidth]{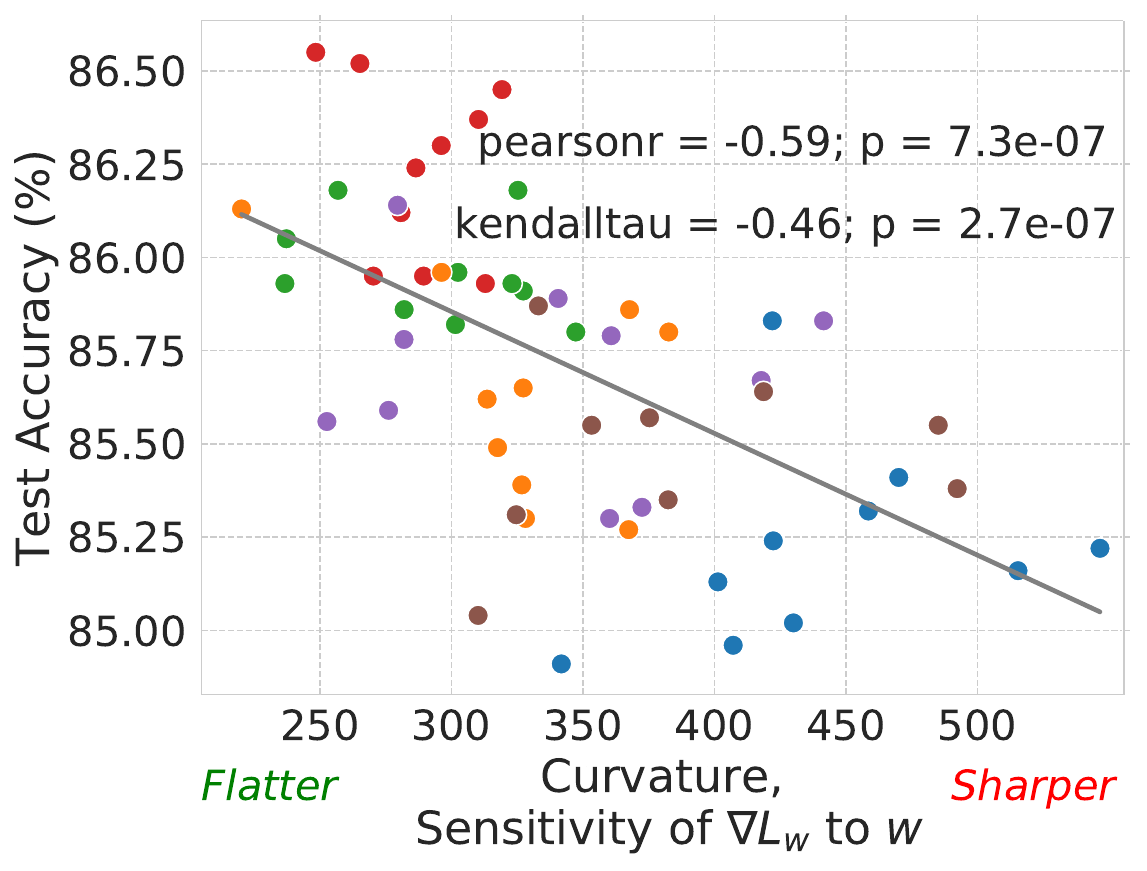}}
    \end{minipage}
    \begin{minipage}[!t]{.4\linewidth}
      \centering
     \centerline{\includegraphics[width=\linewidth]{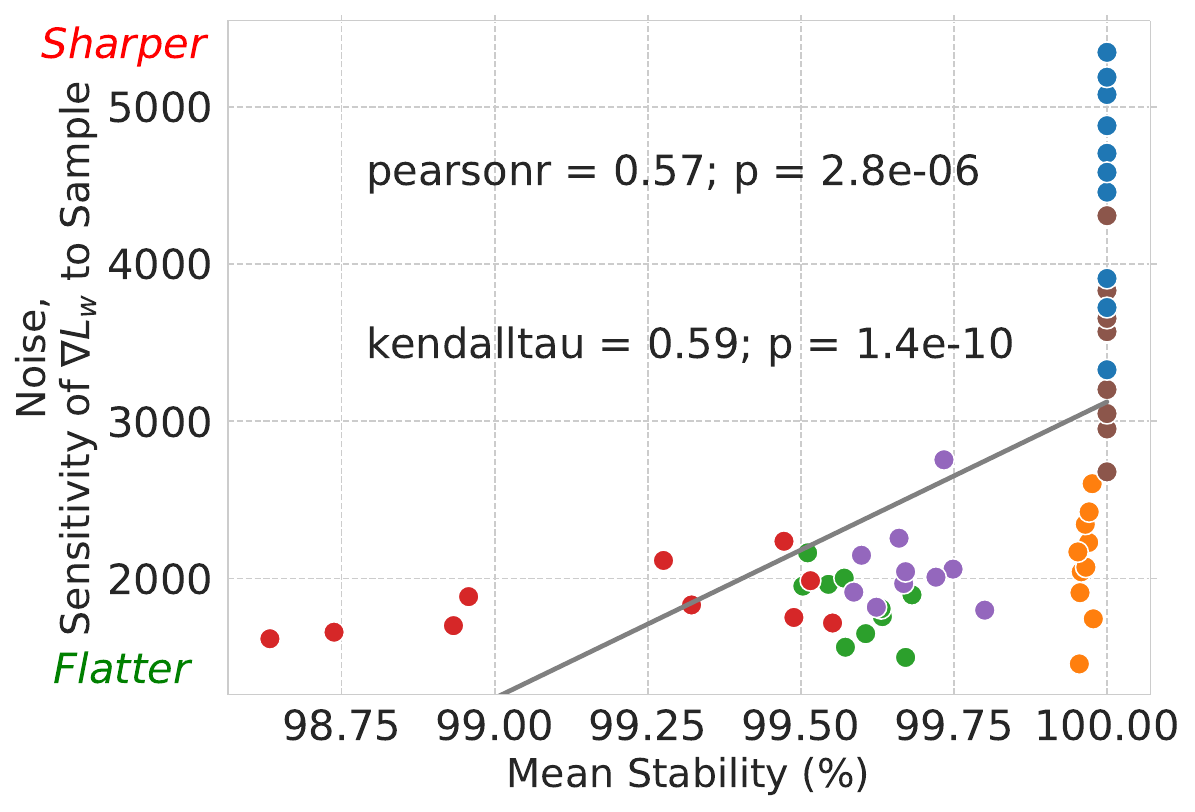}}
    \end{minipage}
    \begin{minipage}[!t]{.09\linewidth}
    \ 
    \end{minipage}
    \begin{minipage}[!t]{.4\linewidth}
      \centering
        \centerline{\includegraphics[width=\linewidth]{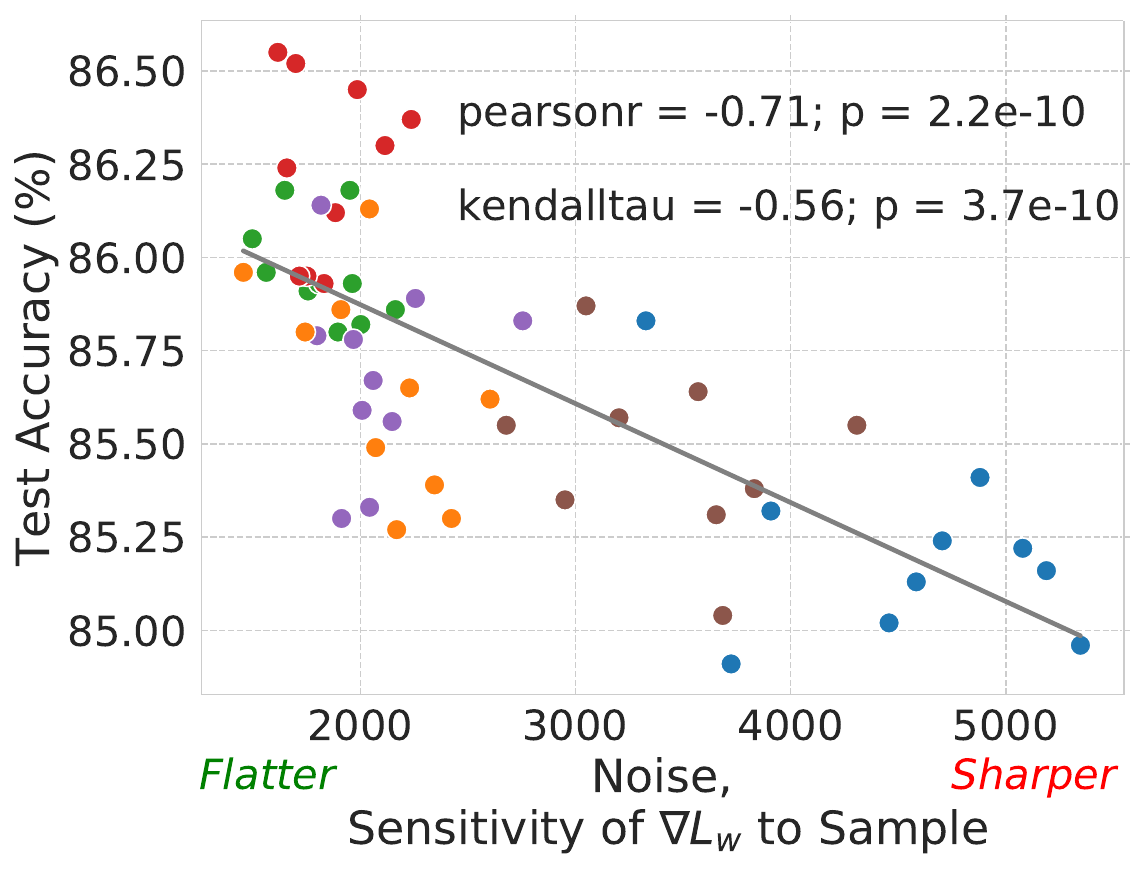}}
    \end{minipage}
  \centering
    \begin{minipage}[h]{.99\linewidth}
      \centering
        \centerline{\includegraphics[width=\linewidth]{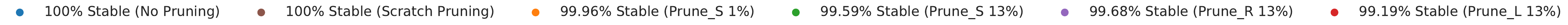}}
    \end{minipage}
\caption{Less pruning stability improves measures of model robustness to noise in the parameters and change in the inputs. These two types of model ``flatness'' are in turn correlated with generalization. ``Scratch'' pruning \cite{li2016pruning,liu2018rethinking} trains the pruned architecture from the outset and is thus 100\% stable.
}
\label{fig:flatness}
\end{figure}

To test these hypotheses, we compute several measures of flatness, and examine their relationships to pruning stability and final generalization in VGG11. We find that there is also a tradeoff between flatness and stability, as decreasing stability led to flatter minima for all flatness measures (Figure \ref{fig:flatness} and Appendix \ref{sec:app_flat}). Furthermore, increased flatness statistically significantly improved generalization. Thus, we find evidence supporting the hypothesis that less pruning stability leads to greater flatness of a kind that is helpful to generalization. 

This result also suggests that the generalization-stability tradeoff we observe may be mediated by increases to the flatness of the converged solution. Specifically, after DNNs recover from the corruption of representations issued by pruning, they not only generalize better but also are less sensitive to data sample and parameter changes. Supporting treatment of pruning as noise injection, this flattening effect is enhanced by representation corruption that is more salient (less stable pruning).

More broadly, these findings add to the recent empirical evidence showing that flatness can explain generalization levels in DNNs when parameter counts cannot \cite{keskar2016large,yao2018hessian}. We also corroborate the recent observation that there is utility in moving beyond parameter flatness and also looking at the gradient covariance to understand generalization performance  \cite{thomas2019interplay,jiang2019fantastic}. Finally, these findings resolve the contradiction between the observation that pruning improves generalization and the emerging generalization theory that de-emphasizes or removes the role of parameter counts \cite{neyshabur2014search,golowich2017size,neyshabur2018towards}. Appendix \ref{sec:app_flat} contains all of our flatness results and details on our measurements of flatness.

\section{Related work}
\label{sec:related}

Many pruning studies have shown that the pruned DNN has heightened generalization \citep{han2015learning,wen2016learning,narang2017exploring,liu2017learning,louizos2017learning,molchanov2017variational,frankle2018lottery,dai2018compressing,ye2018rethinking,you2019gate}, and this is consistent with the fact that pruning may be framed as a regularization (rather than compression) approach. For example, variational Bayesian approaches to pruning via sparsity-inducing priors \citep{molchanov2017variational, louizos2017bayesian} frame weight removal as a means to reduce model description length, which may improve the likelihood of the model obtaining good generalization \citep{rissanen1978modeling}. However, the relevance of the Bayesian/MDL explanation to the regularization done by variational pruning strategies depends on the choice of prior \cite{hron2017variational}. More importantly, non-Bayesian pruning can improve generalization and even outperform variational approaches \cite{gale2019sparsity}, showing that pruning regularizes in non-Bayesian ways.

Pruning to improve generalization has also been inspired by analyses of VC dimension, a measure of model capacity \citep{lecun1990optimal,hassibi1993second}. Overfitting can be bounded above by an increasing function of VC dimension, which itself often increases with parameter counts, so fewer parameters can lead to a guarantee of less overfitting \citep{shalev2014understanding}. While generalization in some learning environments can be eloquently explained by parameter-count-based bounds, such bounds can be so loose in practice that tightening them by reducing parameter counts does not imply better generalization \citep{dziugaite2017computing}. In fact, generalization in deep neural networks tends to \textit{improve} as model size increases \cite{neyshabur2014search,zhang2016understanding,belkin2018reconciling,neyshabur2018towards,nakkiran2019deep}, suggesting that model-size reduction inadequately describes pruning's DNN regularization mechanism (see Appendix \ref{sec:app_c100}). 

More recent generalization bounds consider how the DNN responds to parameter noise \cite{dziugaite2017computing,neyshabur2017exploring,arora2018stronger,nagarajan2019deterministic}, which (along with the gradient covariance) is predictive of generalization in practice \cite{jiang2019fantastic,thomas2019interplay}. Our results provide empirical support for such theory, as we find that iterative DNN pruning may improve both generalization and flatness by creating various noisy versions of the internal representation of the data, which unpruned parameters try to fit to, as in noise-injection regularization \citep{hinton2012improving,wan2013regularization,srivastava2014dropout, poole2014analyzing}.

Flatness and neural network pruning were previously linked by an algorithm that removed weights when doing so led to a flatter loss surface \cite{hochreiter1997flat}. We show that a flat-minimum-search algorithm is not required to flatten models via pruning: simple magnitude pruning injects noise that flattens DNNs.

Dropout creates particularly similar noise to pruning, as it temporarily sets random subsets of layer outputs to zero (likely changing an input's internal representation every epoch). Indeed, applying dropout-like zeroing noise to a subset of features during training can encourage robustness to a post-hoc pruning of that subset \citep{leclerc2018smallify,gomez2018targeted}. The iterative DNN pruning noise analyzed in our experiments differs, however, as it is: applied less frequently, permanent, not random, and less well studied.

When pruning noise is strong enough to alter DNN predictions, accuracy will likely move closer to chance-level, in which case we say the pruning \textit{stability} (defined in Figure \ref{fig:timeline}) falls. The pruning literature has other measures of pruning's impact on the network, including how much pruning affects the values of the weights in the resulting subnetwork (the unpruned weights) via the Euclidean distance between two subnetwork copies trained with and without the removal of the weights targeted by pruning \cite{frankle2019lottery}. Our stability measure characterizes an immediate change in accuracy caused by pruning, allowing us to study how noise injection relates to pruning's effect on generalization.

Permanent removal of parameters is not required to obtain generalization benefits of pruning with DSD (dense-sparse-dense training), retraining a model after pruning then returning the pruned weights to the model for a final training phase \cite{han2016dsd}. Relative to DSD, we demonstrate the effects of multiple different pruning schemes and argue that a scheme with less stability produces better generalization.

\section{Discussion}
\label{sec:disc}

We demonstrated the presence of a generalization-stability tradeoff in neural network pruning that stems from the generalization benefits of pruning less stably, which heightens flatness by intensifying a noise-injection-like effect that does not require permanent parameter removal to be effective. Thus, our results show how pruning-based generalization improvements can be consistent with generalization bounds that do not depend on parameter counts \cite{golowich2017size,neyshabur2018towards}, and they provide empirical support for generalization theory based on flatness/noise-robustness \cite{neyshabur2017exploring,dziugaite2017computing,arora2018stronger}.

Our results suggest that the generalization-stability tradeoff is a useful framework for analyzing the effect of pruning hyperparameters on pruned-model generalization. For example, the fact that iterative pruning outperforms one-shot pruning \cite{han2015learning} can be seen through this framework as an observation about repeated noise injections being preferable to one (perhaps unhelpfully large) injection of noise.

\section*{Broader Impact}

This work focuses on resolving an apparent contradiction in the scientific understanding of the relationship between pruning and generalization performance. As such, we believe its primary impact will be on other researchers and it is unlikely to have substantial broader impacts. That said, understanding the mechanisms underlying our models is important for the safe deployment of such models in application domains. Our work takes a step in that direction, and we hope may help pave the way for further understanding.

\begin{ack}
We thank Juan Guillermo Llanos, Margaret Scheiner, Valentin Thomas, and our reviewers for helpful conversations and feedback.
\end{ack}

\bibliographystyle{unsrtnat} 
\bibliography{bibliography}

\clearpage

\appendix
\counterwithin{figure}{section}
\counterwithin{table}{section}

\section{Experimental design and notation} 
\label{sec:app_cfg}

Here we provide additional details helpful for reproducing our experiments, which are available at \url{https://github.com/bbartoldson/GeneralizationStabilityTradeoff/}.

\subsection{Training environment}
\label{sec:details}

All models were trained with an Ubuntu or Red Hat OS; PyTorch version $=1.4$ \citep{paszke2017automatic}; a single GTX 1080, GTX 1080 Ti, or TITAN X GPU; and CUDA version $10$ or $10.1$. We used Nvidia drivers 435.21, 440.33.01, 450.51.06. For a currently unknown reason, after a computer had its driver updated to the beta driver 455.23.05, correlations in some of our results weakened. 

\subsection{Data}
Our experiments in the main text used the CIFAR-10 \citep{krizhevsky2009learning} dataset from the PyTorch torchvision package, with the default training and testing splits. In Appendix \ref{sec:app_c100}, we used CIFAR-100 data and data augmentation to mimic the data used in \cite{nakkiran2019deep}.

\subsection{Optimization and learning rate schedule notation}

Except in Figures \ref{fig:resnetM} and \ref{fig:beat_baseline}, we trained models using Adam \citep{kingma2014adam} with initial learning rate 0.001. We found Adam more helpful than SGD for recovering accuracy after pruning (perhaps related to the observation  that recovery from pruning is more difficult when learning rates are low \cite{zhu2017prune}). In Figures \ref{fig:resnetM} and \ref{fig:beat_baseline}, we used SGD with the learning rate settings and schedule used in \cite{he2016deep}.

\paragraph{Learning rate schedule notation} In the following experimental details, we specify usage of a multi-step learning rate schedule with $lr_s = (x, y)$, which means we shrink the learning rate by a factor of 10 at epochs $x$ and $y$.

\subsection{Pruning approach and notation}

The hyperparameter settings of our pruning approaches are given using a notation that we describe here. The specific hyperparameter settings used in an experiment are found in the experiment's corresponding appendix section, but several pruning hyperparameter/approach settings are applicable to all of our experiments, and we describe them and how we implement pruning below.

\paragraph{Pruning hyperparameter notation}
We denote the pruning of $n$ layers of a network by specifying a series of epochs at which pruning starts $s=(s_1,...,s_n)$, a series of epochs at which pruning ends $e=(e_1,...,e_n)$, a series of fractions of parameters to remove $p=(p_1,...,p_n)$, and an inter-pruning-iteration retrain period $r \in \mathbb{N}$. For a given layer $l$, the retrain period $r$ and fraction $p_l$ jointly determine the iterative pruning percentage $i_l$. Our experiments prune the same number of parameters $i_l \times \mathrm{size}(layer_l)$ per pruning iteration, ultimately removing $p_l \times 100$\% of the parameters by the end of epoch $e_l$.  

When layerwise iterative pruning percentages differ (i.e., when there exists an $a$ and $b$ such that $i_a$ and $i_b$ are unequal), our plots state the largest iterative pruning rate that was used in any of the layers.

\paragraph{General pruning approach}

We use structured magnitude pruning techniques, i.e., we remove entire filters \cite{li2016pruning}. To score filters for VGG11, we use the $\ell_2$-norm of the filter weights, which was found to perform similarly to  $\ell_1$-norm scoring in \cite{li2016pruning}. Except in Figures \ref{fig:resnetM} and \ref{fig:tradeoff_bn}, we score filters in ResNet18 via the average $\ell_1$-norm of their corresponding feature map activations\footnote{We use an exponential moving average that weights the prior average by $0.9$ and is updated every ten batches during training; \cite{li2016pruning} computes this average at one point in time and thus using a constant set of weights.} \cite{polyak2015channel,li2016pruning} before the non-linear activation but \textit{after} batch-normalization, which enables us to account for incoming shortcut connections (that are added \textit{after} batch-normalization) when judging output-feature-map importance. Relatedly, for ResNets, in addition to pruning the filter associated with a targeted output-feature-map, we also prune any shortcut connections to said map, ensuring its total removal. All models shown in Figure \ref{fig:resnetM} (ResNet20, ResNet56, and ResNet18) were pruned with $\ell_1$-norm scoring of filters. Finally, we studied our own filter scoring methods, discussed in Appendix \ref{sec:app_trade}. 

Given the starting epoch $s_i$, ending epoch $e_i$, retrain period $r$, and fraction to remove $p_i$ for layer $i$, we run a pruning iteration every $r$ epochs, leading to $n_\mathrm{iter}=\floor{\frac{e_i-s_i+1}{r}}+1$ pruning iterations indexed by $j \in (1,n_\mathrm{iter})$. Assuming the layer has $n_\mathrm{filter}$ filters, the ultimate number of remaining filters will be $m_i = \ceil{(1-p_i) \times n_\mathrm{filter}}$. The number of filters that will remain after pruning iteration $j$ is $\ceil{m_i + 
        (n_\mathrm{filter} - m_i) \times (n_\mathrm{iter}-j)/(n_\mathrm{iter})}$, and the number of parameters pruned on iteration $j$ is set accordingly.

To study how pruning-based generalization improvements are affected by reductions in stability from a starting point of \textit{high} stability, we skew pruning toward later layers to allow relatively high stability with all of our various pruning targets (networks were found to be more resilient to pruning of later layers in \cite{li2016pruning,you2019gate}, and we observed a similar pattern when we briefly tried pruning more from earlier layers). One of the ways that we study how pruning-based generalization improvements are affected by reductions in stability from a starting point of \textit{low} stability is raising total pruning percentage. As we discuss in  Section \ref{sec:iter} and  Appendix \ref{sec:app_hyper}, this higher total percentage causes Prune$_\mathrm{L}$ to be much less stable and to perform worse than Prune$_\mathrm{S}$, replicating the observation that Prune$_\mathrm{S}$ outperforms Prune$_\mathrm{L}$ at higher total pruning percentages \cite{li2016pruning}.

\section{The generalization-stability tradeoff experiment configuration and results with other scoring methods}
\label{sec:app_trade}

\subsection{Figure \ref{fig:tradeoff_l2} configuration}
\label{sec:details_corr}

The models were trained on CIFAR-10 with Adam for 325 epochs with $lr_s = (150,300)$. The error bars are 95\% confidence intervals for the mean, bootstrapped from 10 distinct runs of each experiment.

\paragraph{VGG11}
We pruned the final four convolutional layers during training with (layerwise) starting epochs $s=(3,4,5,6)$, ending epochs $e=(150,150,150,275)$, and pruning fractions $p=(0.3,0.3,0.3,0.9)$. To allow for the same amount of pruning among models with differing iterative pruning percentages, we adjusted the number of inter-pruning retraining epochs. The models with the maximum iterative pruning percentage of 1\% had $r=4$, while the models with the maximum iterative pruning percentage of 13\% had $r=40$. 

\paragraph{ResNet18}
We pruned the final four convolutional layers of ResNet18 during training with (layerwise) starting epochs $s=(7,8,9,10)$, ending epochs $e=(150,150,170,275)$, and pruning fractions $p=(0.25,0.4,0.25,0.95)$.  The models with the maximum iterative pruning percentage of 1\% had $r=4$, while the models with the maximum iterative pruning percentage of 14\% had $r=40$.

While we could lower pruning stability in ResNet18, this model interestingly adapted to pruning events much differently than VGG11: test accuracy rebounded after pruning as in VGG11 but then quickly flattened out rather than climbing steadily as the network adapted to the pruning, which may be related to noise stability properties created by shortcut connections \cite{yu2019identity}. Thinking that shortcut connections were allowing the network to adapt to pruning events too easily, we tried pruning a larger amount  of the penultimate block's output layer (moving from 0.25 to the shown 0.4), which reduced the number of shortcut connections to the final block's output layer, lengthened the adaptation period, and increased pruning-based generalization improvements.

\subsection{Changing the scoring method used in Figure \ref{fig:tradeoff_l2}}
\label{sec:new_score}

We found that we could strengthen the correlations shown in Figure \ref{fig:tradeoff_l2} by switching the pruning scoring method from an $\ell_2$-norm approach to one that made stable approaches more stable, and unstable approaches more unstable (reducing the effect of measurement noise on the correlation). The correlations strengthened from those shown in Figure \ref{fig:tradeoff_l2} to -0.88 and -0.65 for VGG11 and ResNet18, respectively. Here we describe this pruning scoring method and the results.

\subsubsection{A scoring method to identify important batch-normalized parameters}
\label{sec:e[bn]}

The correlation between a parameter's magnitude and its importance-to-the-loss weakens in the presence of batch normalization (BN) \citep{ioffe2015batch}. Without batch normalization, a convolutional filter with weights $W$ will produce feature map activations with half the magnitude of a filter with weights $2W$: filter magnitude clearly scales the output. With batch normalization, however, the feature maps are normalized to have zero mean and unit variance, and their ultimate magnitudes depend on the BN affine-transformation parameters $\gamma$ and $\beta$. As a result, in batch normalized networks, filter magnitude does not scale the output, and equating small magnitude and unimportance may therefore be particularly flawed. This has motivated approaches to use the scale parameter $\gamma$'s magnitude to find the convolutional filters that are important to the network's output \citep{ye2018rethinking,you2019gate}.  Here, we derive a novel approach to determining filter importance/magnitude that incorporates both $\gamma$ and $\beta$.

To approximate the expected value/magnitude of a batch-normalized, post-ReLU feature map activation, we first define the 2D feature map produced by convolution with BN:

\vspace{-4mm}
$$
M =  \gamma \mathrm{BN}(W*x) + \beta.
$$
\vspace{-4mm}

We approximate the activations within this feature map as $M_{ij} \sim \mathcal{N}(\beta, \gamma')$, where $\gamma'=\left| \gamma \right|$. This approximation is justified by the central limit theorem when the products in $W*x$ are i.i.d. and sufficiently numerous; empirically, we show in Figure \ref{fig:normality} that this approximation is highly accurate early in training but becomes less accurate as training progresses. Given this approximation, the post-ReLU feature map

\vspace{-4mm}
$$
R = \max \{ 0, M \}
$$
has elements $R_{ij}$ that are either $0$ or samples from a truncated normal distribution with left truncation point $l=0$, right truncation point $r=\infty$, and mean $\mu$ where

\vspace{-4mm}
\begin{gather*}
 \mu = \gamma' \frac{\phi(\lambda)-\phi(\rho)}{Z} + \beta, \\[3pt]
 \lambda=\frac{l-\beta}{\gamma'}, \rho=\frac{r-\beta}{\gamma'},  Z=\Phi(\rho)-\Phi(\lambda),
\end{gather*}
and $\phi(x)$ and $\Phi(x)$ are the standard normal distribution's PDF and CDF (respectively) evaluated at $x$. Thus, an approximation to the expected value of $R_{ij}$ is given by

\vspace{-4mm}
$$
\mathbb{E}[R_{ij}] \approx  \Phi(\lambda)0 + (1-\Phi(\lambda)) \mu.
$$

We use the phrase "\textit{E[BN] pruning}" to denote magnitude pruning that computes filter magnitude using this derived estimate of $\mathbb{E}[R_{ij}]$. E[BN] pruning has two advantages. First, this approach avoids the problematic assumption that filter importance is tied to filter $\ell_2$ norm in a batch-normalized network. Accordingly, we hypothesize that E[BN] pruning can grant better control of the stability of the neural network's output than pruning based on filters' $\ell_2$ norms. Second, the complexity of the calculation is negligible as it requires (per filter) just a few arithmetic operations on scalars, and two PDF and CDF evaluations, making it cheaper than approximating the expected value via the sample mean of feature map activations for a batch of feature maps.

\subsubsection{Quality of normality approximation by layer and training level}
\label{sec:derivation}

\begin{figure}[h]
    \centering
     \centerline{\includegraphics[width=.4\linewidth]{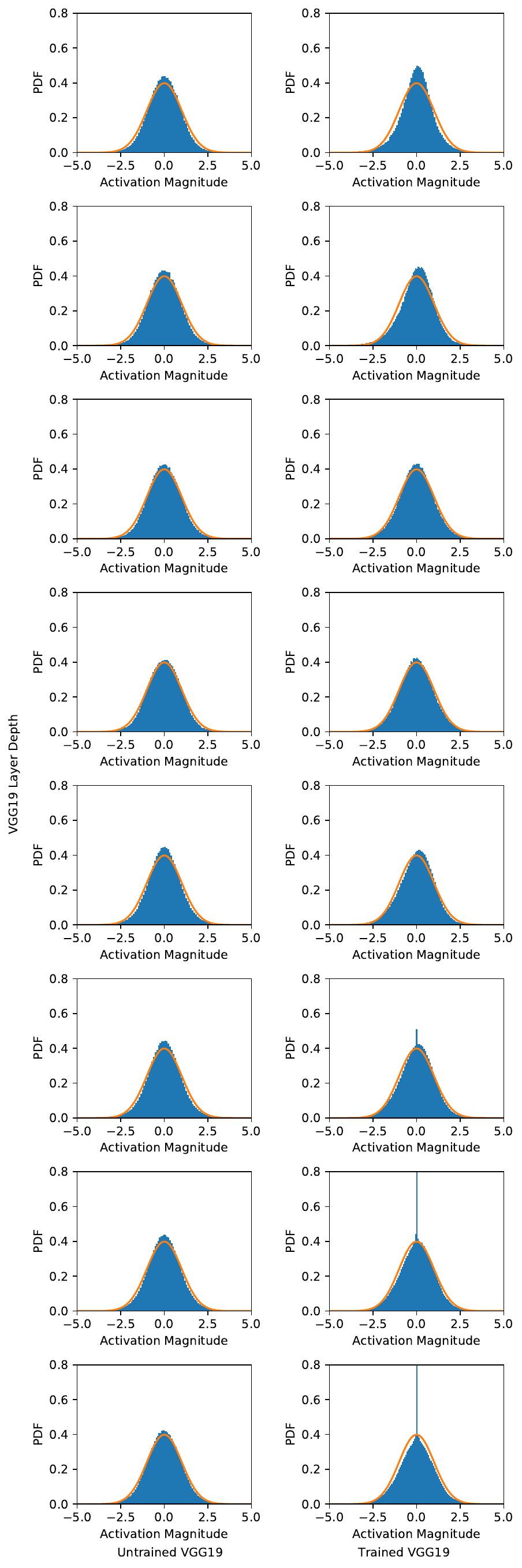}}
    \caption{We examined the normalized activations (shown in blue histograms) of feature maps in the final eight convolutional layers of VGG19 before training (left) and after training (right). We found that the approximation to standard normality (shown in orange) of these activations is reasonable early on but degrades with training (particularly in layers near the output).}
    \label{fig:normality}
\end{figure}

The main drawback to the E[BN] approach (Section \ref{sec:e[bn]}) is the sometimes poor approximation $M_{ij} \sim N(\beta, \gamma')$, which depends on the assumption of $N(0, 1)$ distributed  feature map activations (after batch normalization, but before the associated affine transformation). In Figure \ref{fig:normality}, this assumption's validity depends on layer and the training of the model (we used a pre-trained model from the PyTorch torchvision package to show the effect of the latter on the approximation). A less serious drawback is that this approach does not account for the strength of connections to the post-BN feature map, which could have activations with a large expected value but low importance if relatively small-magnitude weights connected the map to the following layer.

\subsubsection{Updating figure \ref{fig:tradeoff_l2} with new scoring methods}
\label{sec:tradeoff_bn}

\begin{figure}[!t]
  \centering
    \begin{minipage}[h]{.65\linewidth}
      \centering
     \centerline{\includegraphics[width=\linewidth]{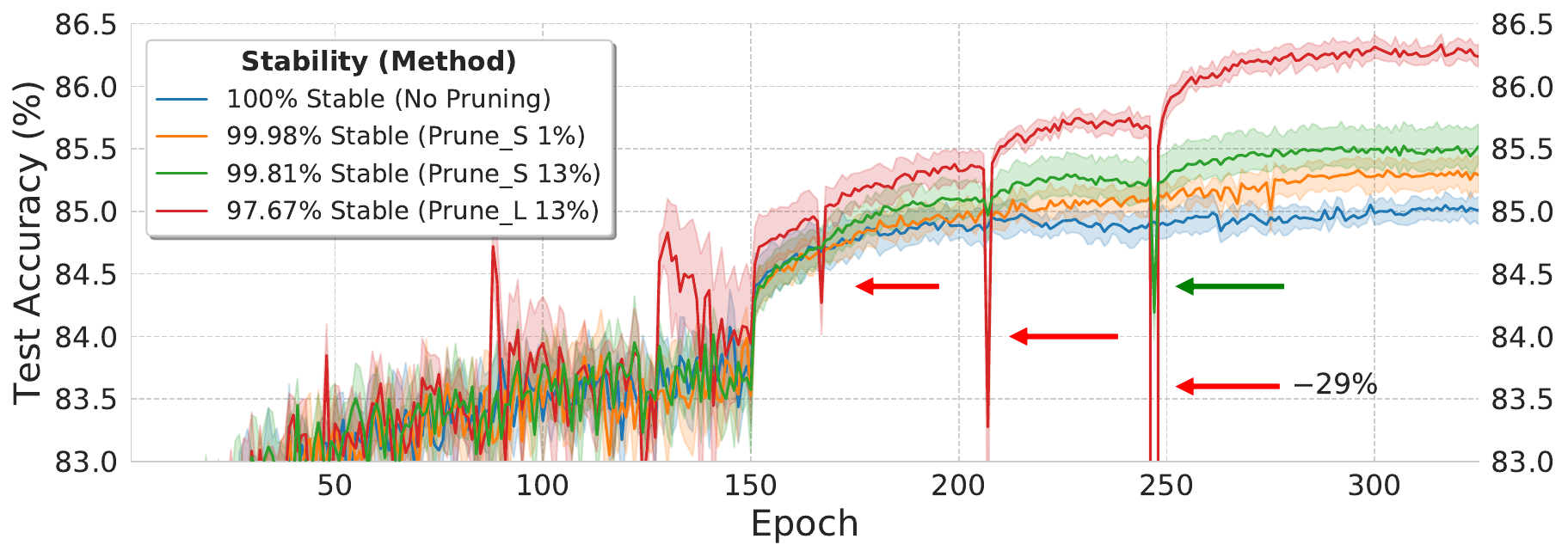}}
    \end{minipage}
    \begin{minipage}[!t]{.34\linewidth}
      \centering
        \centerline{\includegraphics[width=\linewidth]{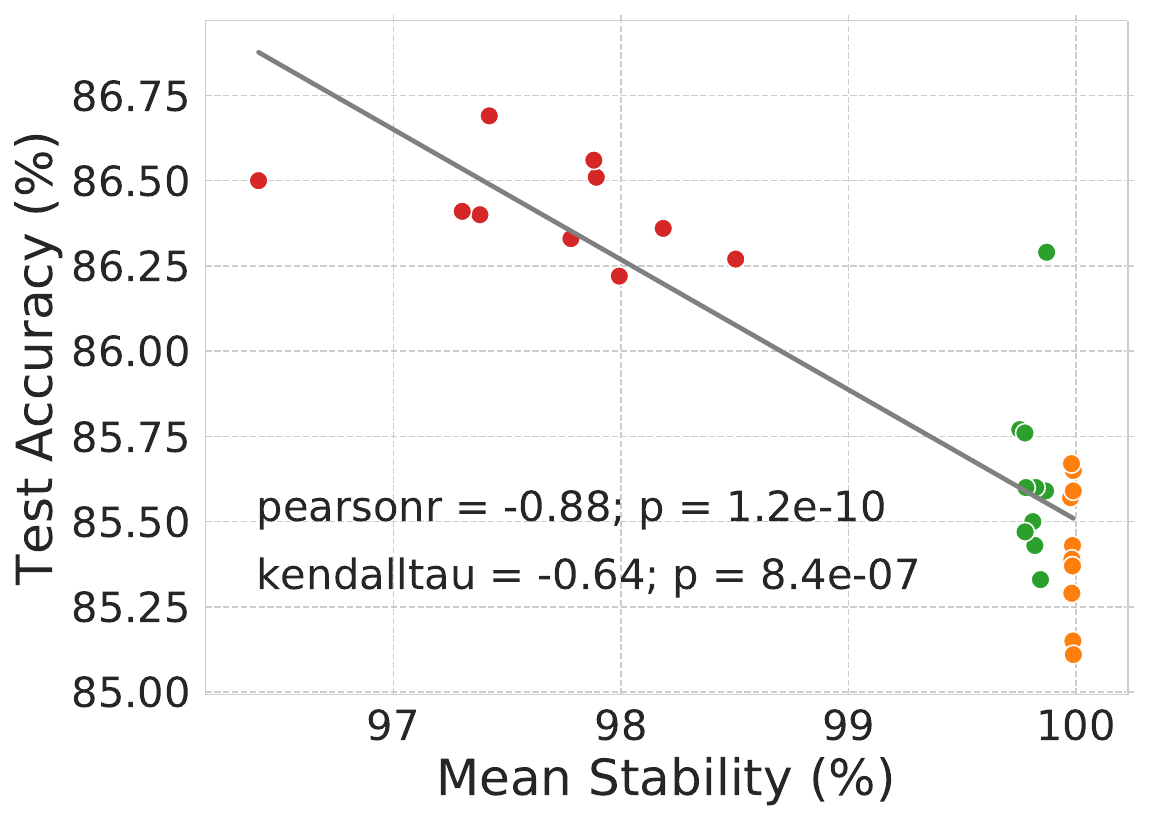}}
    \end{minipage}
    \begin{minipage}[!t]{.65\linewidth}
      \centering
     \centerline{\includegraphics[width=\linewidth]{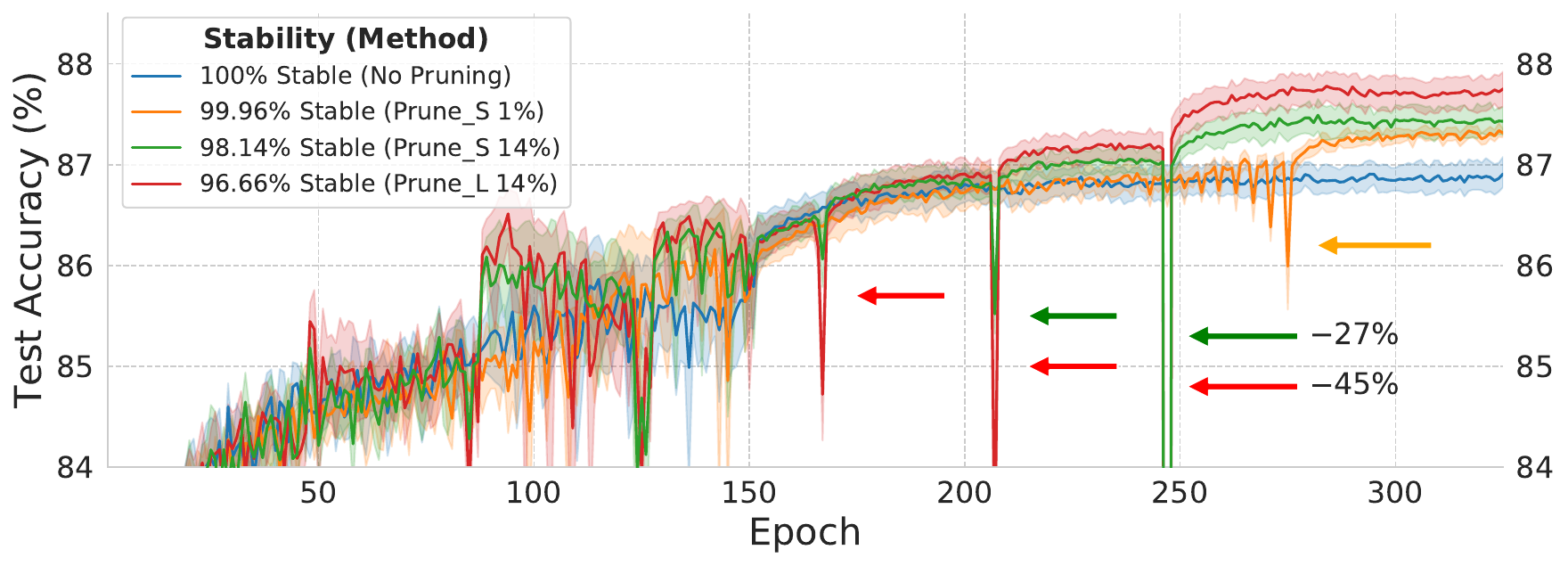}}
    \end{minipage}
    \begin{minipage}[!t]{.34\linewidth}
      \centering
        \centerline{\includegraphics[width=\linewidth]{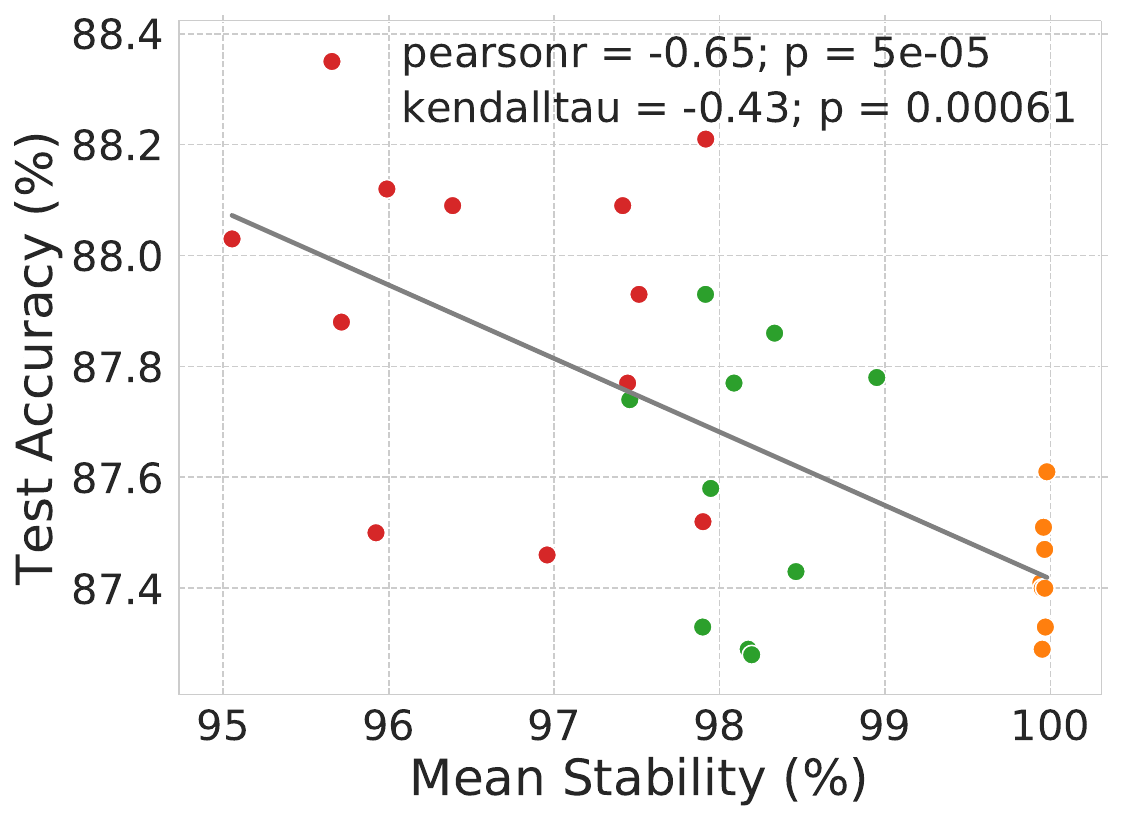}}
    \end{minipage}
\caption{Less pruning stability improves generalization of (Top) VGG11 and (Bottom) ResNet18 when training on CIFAR-10 (10 runs per configuration). (Left) Test accuracy during training of several models illustrates how adaptation to less stable pruning leads to better generalization. (Right) Means reduce along the epoch dimension (creating one point per run-configuration combination).}
\label{fig:tradeoff_bn}
\end{figure}

Figure \ref{fig:tradeoff_bn} shows the results of switching the scoring method used in Figure \ref{fig:tradeoff_l2} to new scoring methods. Figure \ref{fig:tradeoff_bn} uses the same configuration as Figure \ref{fig:tradeoff_l2},  described in Section \ref{sec:details_corr}, except ResNet18 starts pruning slightly earlier with $s=(3,4,5,6)$.

For VGG11, we use the scoring method described in Section \ref{sec:e[bn]}. When using this scoring method in ResNet18, the correlation did not improve, with Prune$_\mathrm{S}\ 1\%$ in particular remaining relatively unstable. As such, for ResNet18,  Figure \ref{fig:tradeoff_bn} shows a scoring method that is a modification of the scoring method in Section \ref{sec:e[bn]} that replaces $\gamma'$ with  $\gamma$ (precluding this method's interpretation as an approximation to the post-ReLU expected value). While not having a significant effect on VGG11 (its use would change the correlation coefficient from -0.88 to -0.89), this modification creates visibly more stable pruning for Prune$_\mathrm{S}\ 1\%$ in ResNet18. For example, the final drop in accuracy in Figure \ref{fig:tradeoff_bn} is $\approx 1\%$ compared to the $\approx 2\%$ drop in Figure \ref{fig:tradeoff_l2}; quantitatively, the difference  in mean stability across all pruning events between these two methods is minor ($99.957\%$ stability in Figure \ref{fig:tradeoff_l2} vs. $99.961\%$ stability stability in  Figure \ref{fig:tradeoff_bn}), which suggests that the level of regularization created by pruning may be better explained by looking at mean stability in combination with the number of pruning events that fall below some stability threshold rather than just mean stability.

\section{Limits of the generalization-stability tradeoff}
\label{sec:app_hyper}

\subsection{Figure \ref{fig:retrain} configuration}
\label{sec:details_retrain}

In Figure \ref{fig:retrain}, pruning targeted the final four convolutional layers of VGG11 during training with (layerwise) starting epochs $s=(3,4,5,6)$, ending epochs $e=(150,150,150,275)$, and pruning fractions $p=(0.3,0.3,0.3,0.9)$. To create the different iterative pruning rates, we used models with inter-pruning retrain periods $r=4$, $r=20$, $r=40$, $r=60$, $r=100$, $r=200$, and $r=300$. Since the layerwise pruning percentages varied, pruning required multiple iterative pruning percentages, the largest of which is denoted on the horizontal axis.
 
The models were trained on CIFAR-10 with Adam for 325 epochs with $lr_s = (150,300)$. The error bars are 95\% confidence intervals for the means, bootstrapped from 10 distinct runs of each experiment.

\subsection{Higher total pruning percentage configuration and table}

To examine the effect of lowering stability by increasing total pruning percentage, we raised the total pruning percentage of ResNet18 from 46\% to 59\% by setting $p=(0.25,0.8,0.25,0.95)$, otherwise all training details remained as they were in \ref{sec:details_corr}.\footnote{Note that holding constant $r$ and the pruning schedule while raising the total pruning percentage causes the iterative pruning rate to rise. When we specify the pruning method in Table \ref{table:total} we use the iterative rate from Figure \ref{fig:tradeoff_l2}, though the actual iterative rate is higher for the methods with higher total pruning percentages.} For the 59\% experiments, results are based on just three runs per configuration, rather than the typical ten. Means and standard deviations for stability and test accuracy are tabled below (Table \ref{table:total}), in comparison to their corresponding values from Figure \ref{fig:tradeoff_l2}.

Further supporting the idea of a ``sweet-spot'' in the stability level, we find that the two best test accuracies are at  the same stability level,  $98.9\%$ (Table \ref{table:total}). Interestingly, \textit{this stability level was reached with two different pruning targets}, Prune$_\mathrm{S}$ and  Prune$_\mathrm{L}$. Additionally, the two statistically significant changes that we observe are a reduction in stability from a starting point of low stability harming accuracy (when raising the total pruning percentage for Prune$_\mathrm{L}$), and a reduction in stability from a starting point of high stability improving accuracy (when raising the total pruning percentage for Prune$_\mathrm{S}$).
\\
\begin{table}[h]
\caption{Reducing stability by raising total pruning percentage}
\label{table:total}
    \centering
    \begin{tabular}{cccccc}
\\
        \toprule
        \multirow{1}{*}[-1.5em]{Method} & {} &
         \multicolumn{2}{c}{Test Accuracy} & \multicolumn{2}{c}{Stability}\\
        \cmidrule{3-4} \cmidrule{5-6} \\
         {} & 
        \multirow{3}{*}[2.15em]{\shortstack[c]{Total \\ Pruning \\ Percentage}}  & Mean (\%) & Std. Dev. & Mean (\%) & Std. Dev. \\
        \midrule
        \multirow{2}{*}{Prune$_\mathrm{S}\ 1\%$} & 46\% & 87.65  & 0.26 & 99.957  & 0.006 \\
         & 59\% & 87.52  & 0.11 & 99.935  & 0.012 \\
         \\
        \multirow{2}{*}{Prune$_\mathrm{S}\ 13\%$} & 46\% & 87.72  & 0.18 & 99.470  & 0.203 \\
         & 59\% & \ \ 87.95$^\ast$  & 0.18 & 98.912  & 0.303 \\
         \\
        \multirow{2}{*}{Prune$_\mathrm{L}\ 13\%$} & 46\% & 88.03  & 0.26 & 98.904  & 0.229 \\
         & 59\% & \ \ 87.48$^\ast$  & 0.37 & 98.143  & 0.434 \\
        \bottomrule
    \end{tabular}
    
        \vspace{.05cm}
     {\raggedright \quad \quad \quad \quad \ \ \ {$^\ast$} \text{Statistically significant at $<10\%$ significance level with a two-tailed t-test.} \par}
\end{table}

\subsection{Visualization of data used in Figure \ref{fig:retrain} right}
\label{sec:additional}

\begin{figure}[!h]
    \centering
    \begin{minipage}[!t]{.32\linewidth}
      \centering
  \centerline{\includegraphics[width=\linewidth]{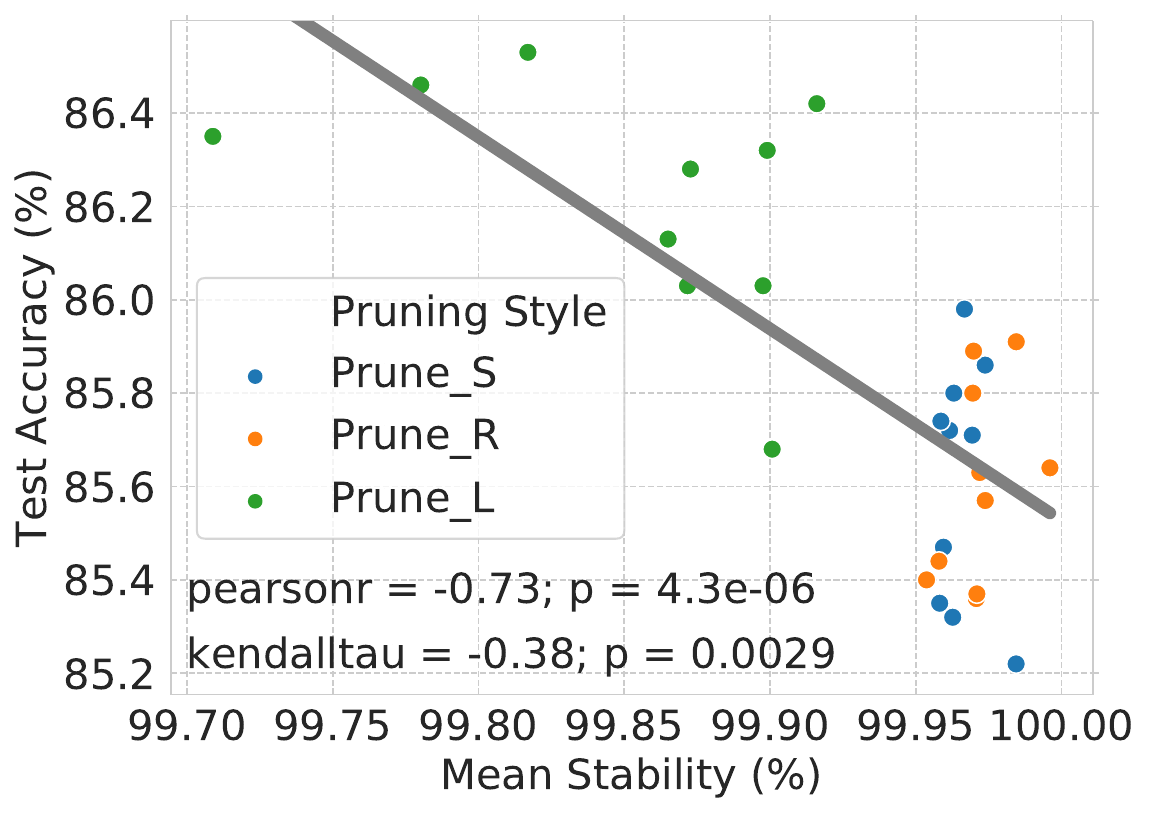}}
    \end{minipage}
    \begin{minipage}[!t]{.32\linewidth}
      \centering
  \centerline{\includegraphics[width=\linewidth]{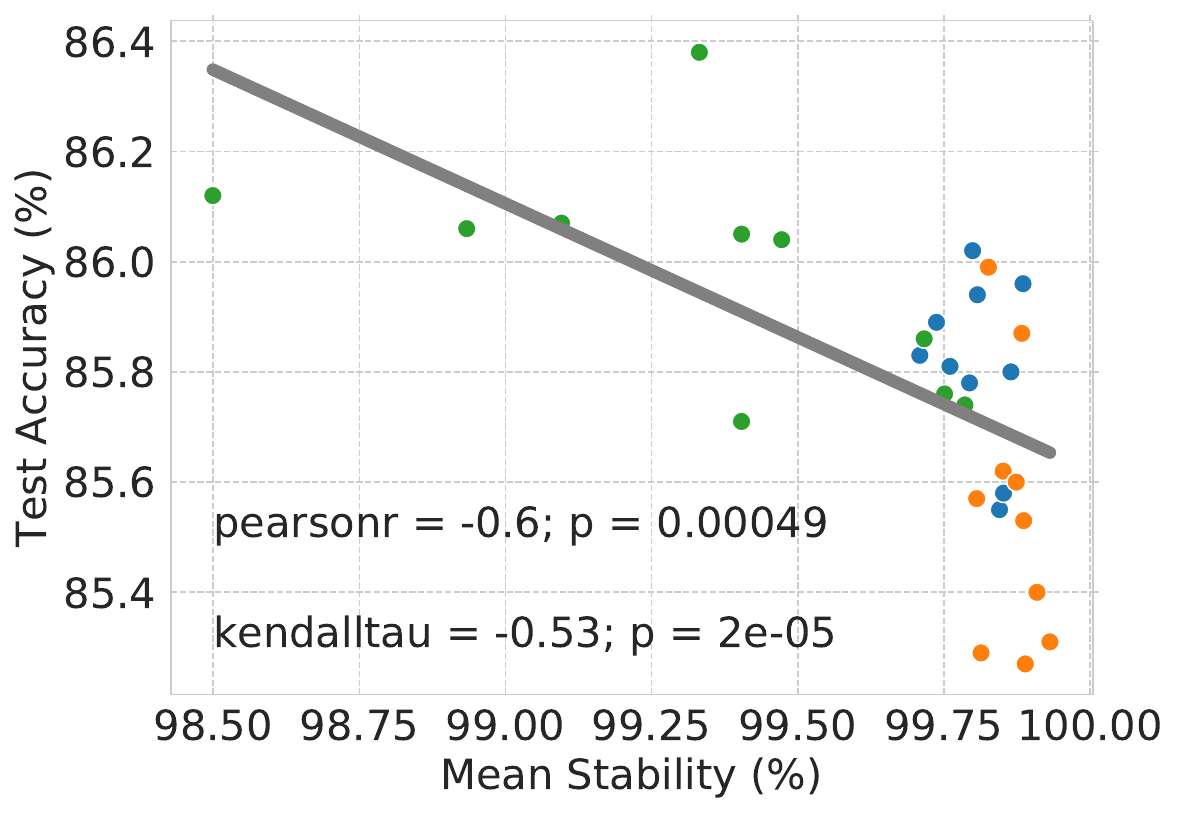}}
    \end{minipage}
    \begin{minipage}[!t]{.32\linewidth}
      \centering
  \centerline{\includegraphics[width=\linewidth]{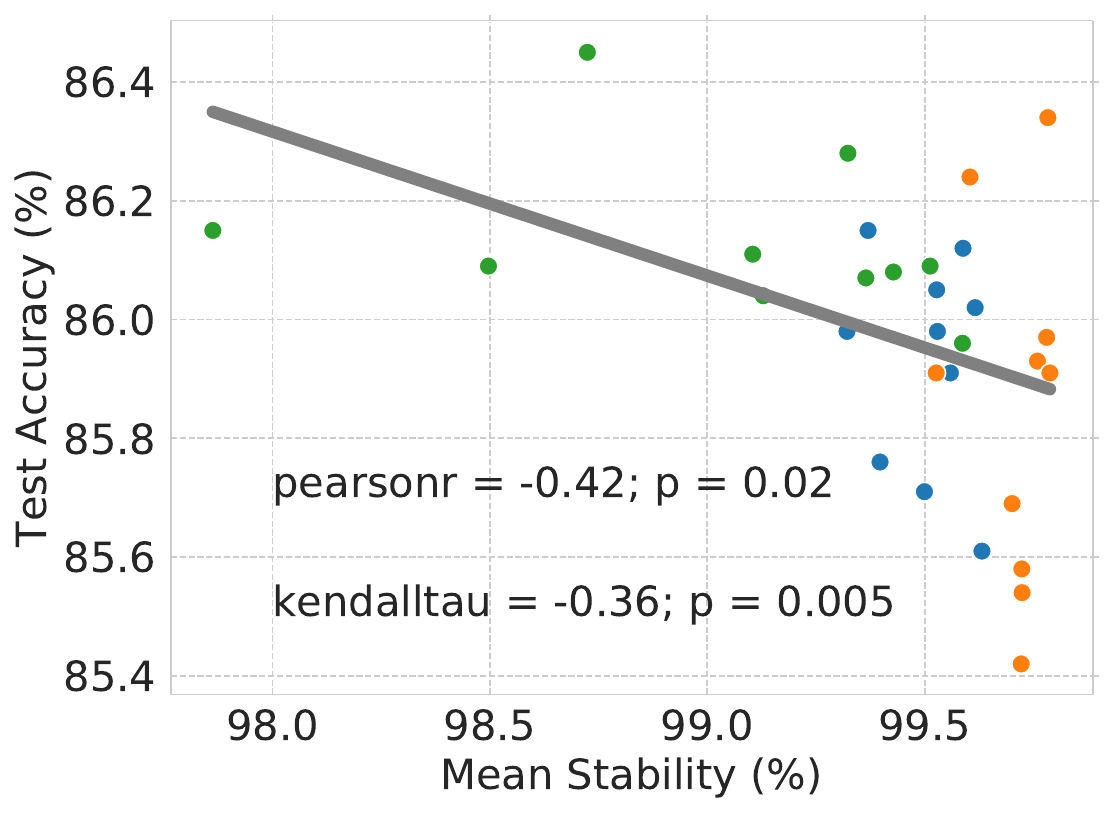}}
    \end{minipage}
    \begin{minipage}[!t]{.32\linewidth}
      \centering
  \centerline{\includegraphics[width=\linewidth]{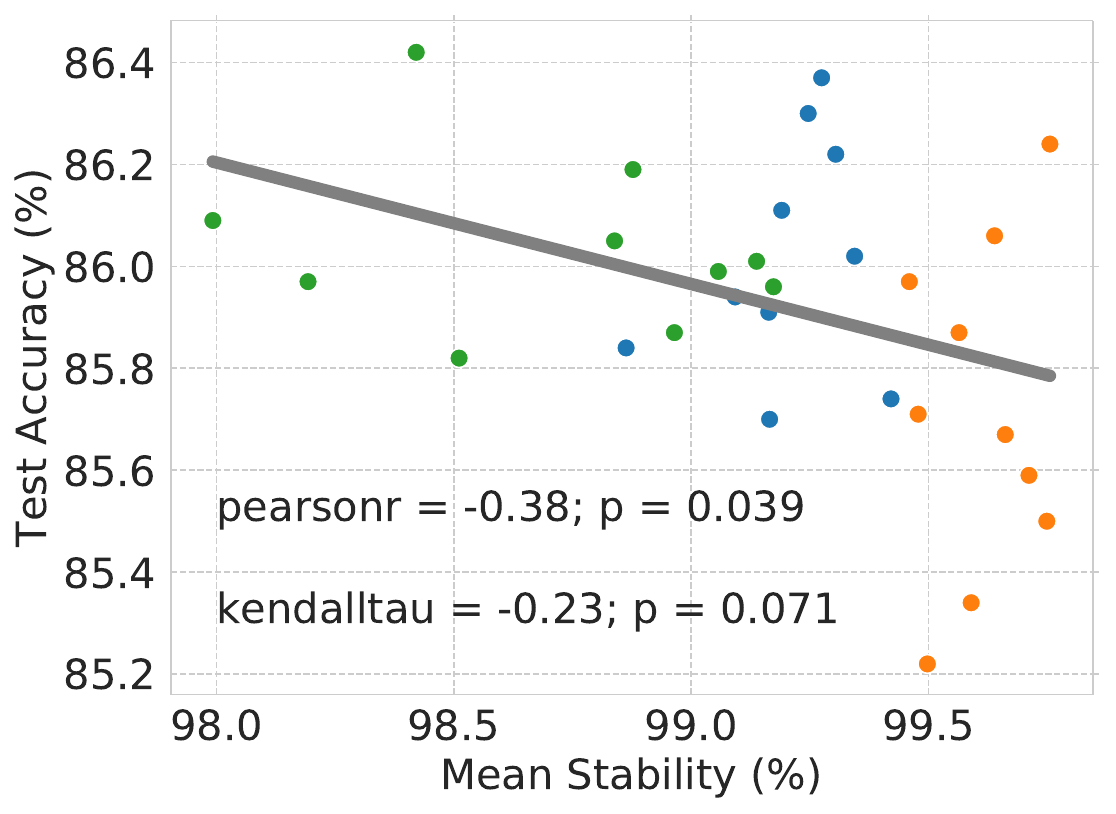}}
    \end{minipage}
    \begin{minipage}[!t]{.32\linewidth}
      \centering
  \centerline{\includegraphics[width=\linewidth]{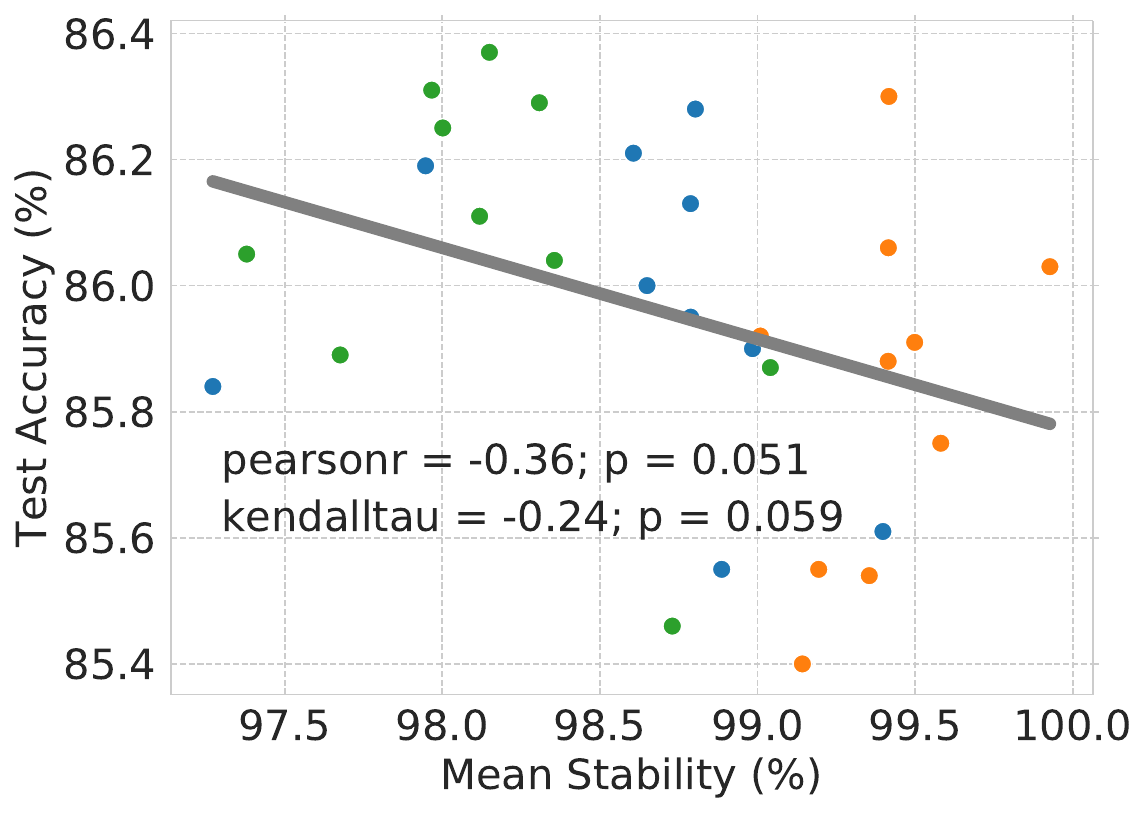}}
    \end{minipage}
    \begin{minipage}[!t]{.32\linewidth}
      \centering
  \centerline{\includegraphics[width=\linewidth]{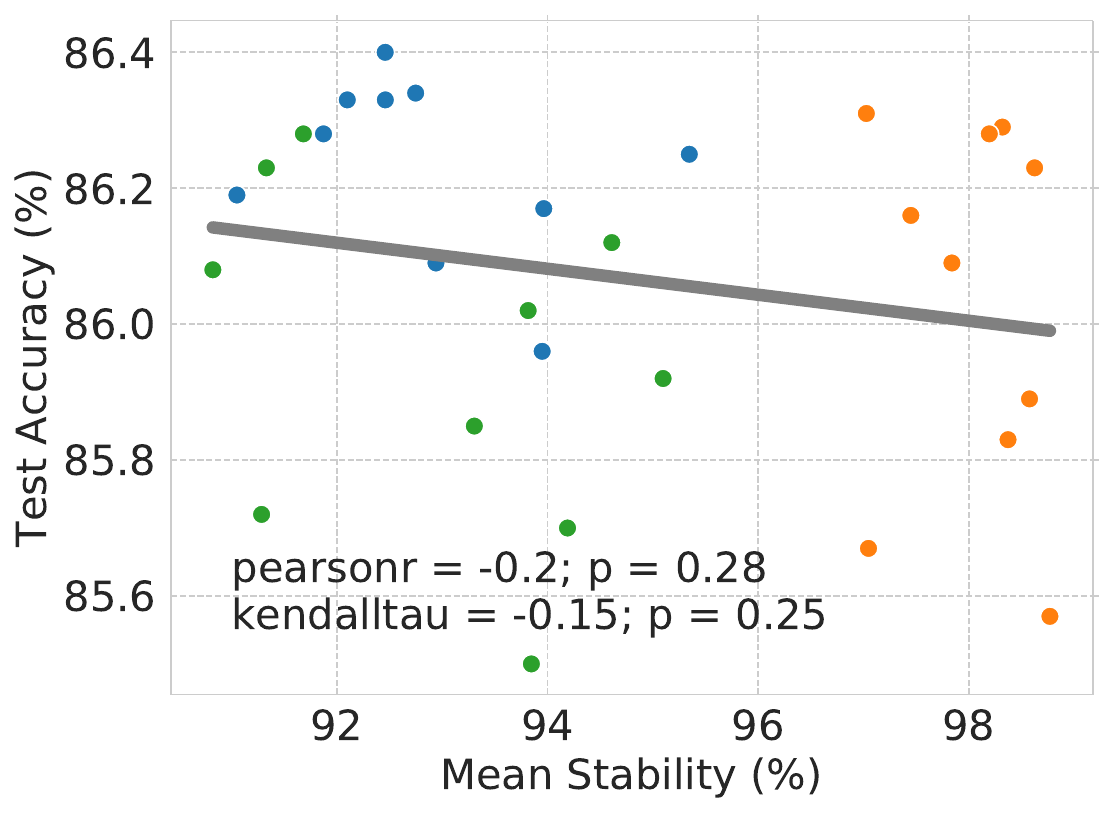}}
    \end{minipage}
    \begin{minipage}[!t]{.32\linewidth}
      \centering
  \centerline{\includegraphics[width=\linewidth]{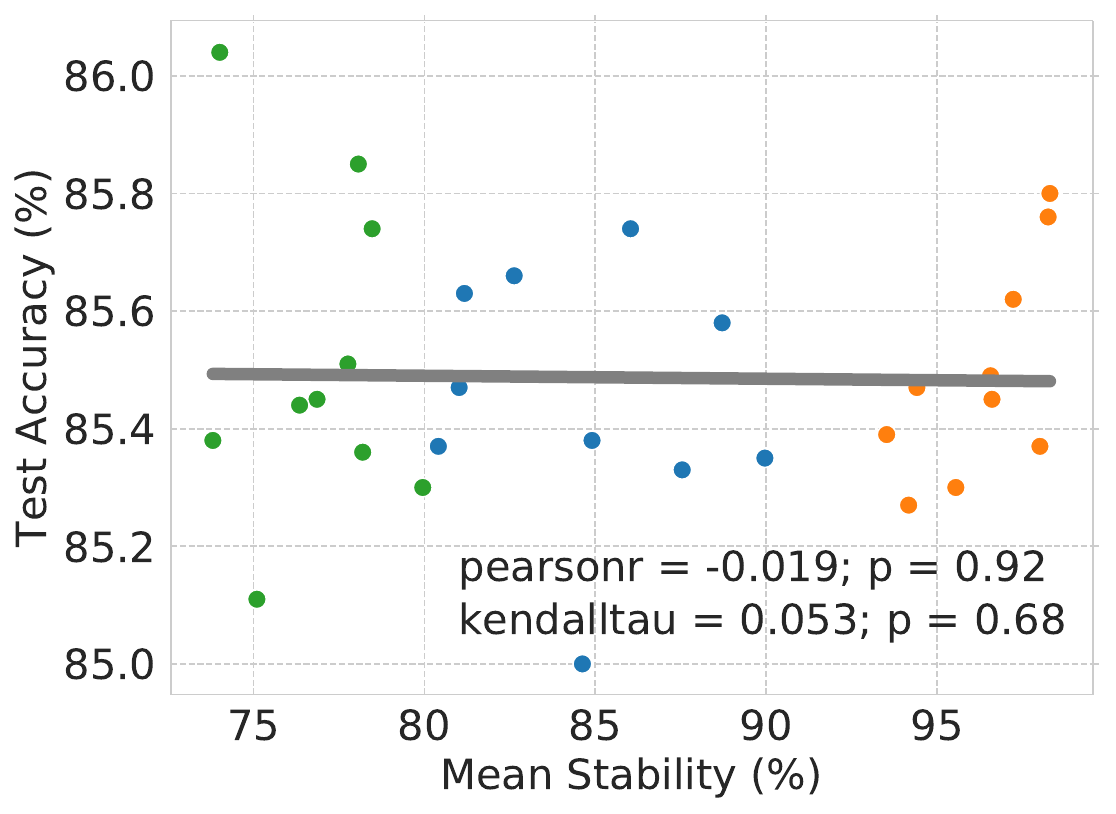}}
    \end{minipage}
\caption{At each iterative pruning rate, reducing pruning stability by targeting more important weights aids generalization. This correlation is typically statistically significantly at the 5\% significance level. Iterative pruning rate increases from left to right, then top to bottom.
}
\label{fig:constant_iter_rate}
\end{figure}

\begin{figure}[!t]
    \centering
    \begin{minipage}[!t]{.4\linewidth}
      \centering
  \centerline{\includegraphics[width=\linewidth]{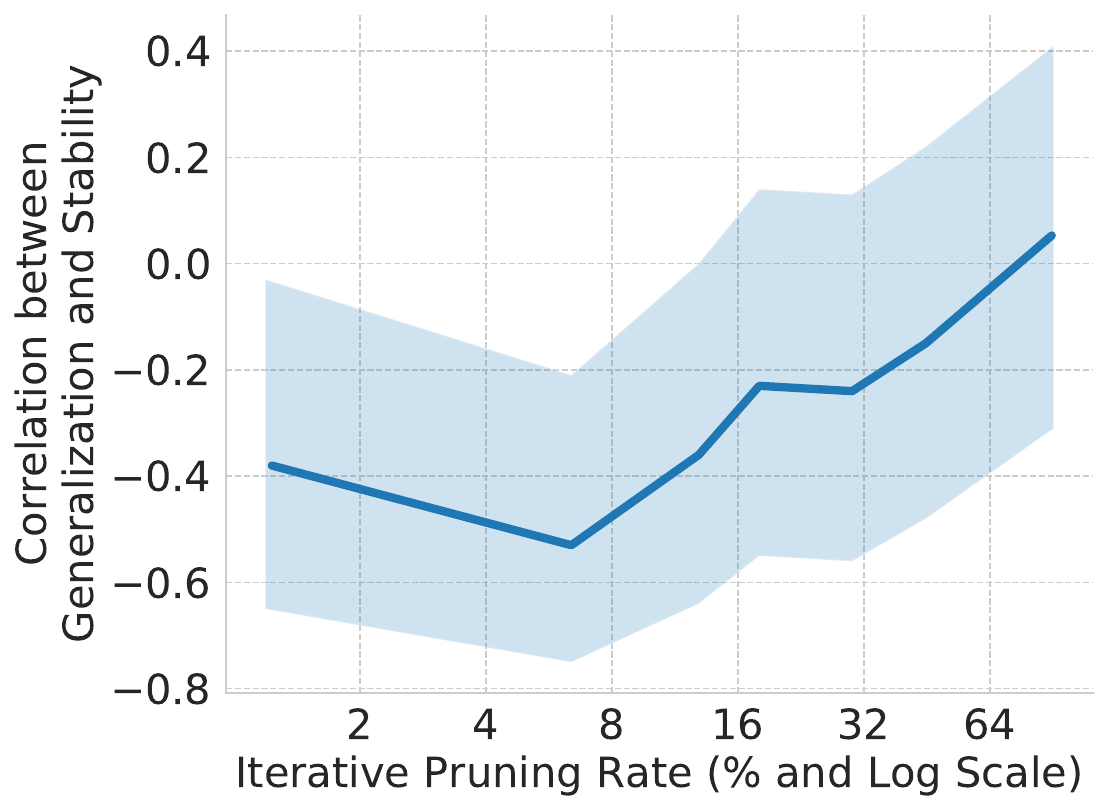}}
    \end{minipage}    
\caption{At a particular iterative rate, Kendall's rank correlation between generalization and stability is always negative, except at the rate corresponding to the one-shot pruning case.}
\label{fig:kendall_corr}
\end{figure}

In Figure \ref{fig:constant_iter_rate}, we raise iterative rate, which reduces pruning stability for all pruning targets. At lower stabilities, the benefit of further decreasing stability by changing target becomes less visible or non-existent. Note that the correlations in Figure \ref{fig:constant_iter_rate} were used to make  the correlation plot in the main text (Figure \ref{fig:retrain} right) and the corresponding summary of the Kendall rank correlations shown in Figure \ref{fig:kendall_corr}.
\vspace{.1cm}

\subsection{More Results with Traditional Setup from Figure \ref{fig:resnetM}}
\label{sec:app_traditional}

In Figure \ref{fig:resnetM} we trained models exactly as specified in \cite{he2016deep}. When doing so, we found that, while the generalization stability tradeoff was present among pruned models, the pruned models didn't outperform the baseline model on average. A possibility is that the pruning procedure may have been too disruptive to be beneficial in the context of the standard training and regularization settings. If this is the case, then our results suggest that there's a higher stability level at which the generalization-stability tradeoff provides a benefit but is not so disruptive that the advantages of the standard training and regularization approach are lost. 

We test this in ResNet18 by switching to a more stable set of pruning schemes. We still prune every layer of every block of ResNet18, but we now prune over a longer period of time, with starting epoch $s=(24)$ and ending epoch $e=(100)$ for each layer, and we consider two pruning fractions for each layer pruned: $p=(0.06)$ and $p=(0.1)$. If the generalization-stability tradeoff can lead to an improvement in a traditionally-trained baseline model's performance when pruning is sufficiently stable, then we would expect to see such pruning methods obtaining higher accuracy than the baseline ResNet18 model, which has 94.35\% test accuracy on average.

\begin{figure}[!t]
    \centering
    \begin{minipage}[!t]{.4\linewidth}
      \centering
  \centerline{\includegraphics[width=\linewidth]{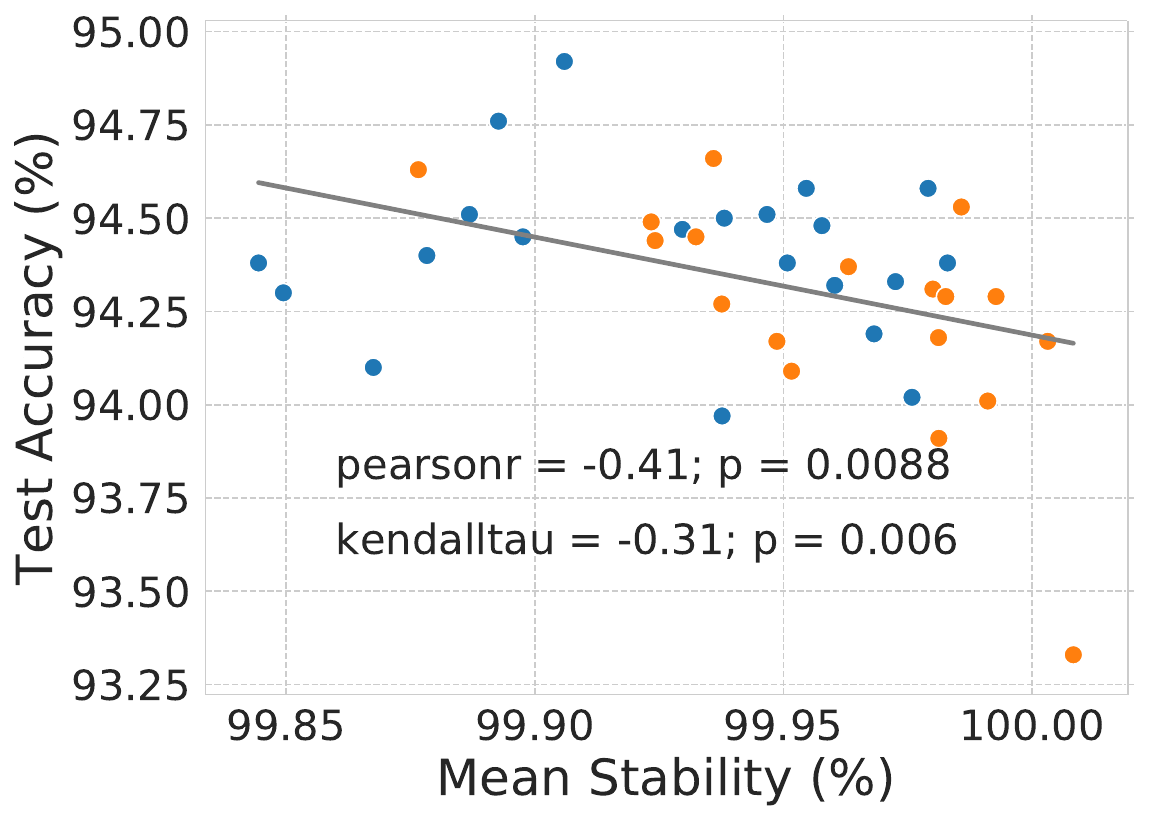}}
    \end{minipage}    
    \begin{minipage}[!t]{.4\linewidth}
      \centering
  \centerline{\includegraphics[width=\linewidth]{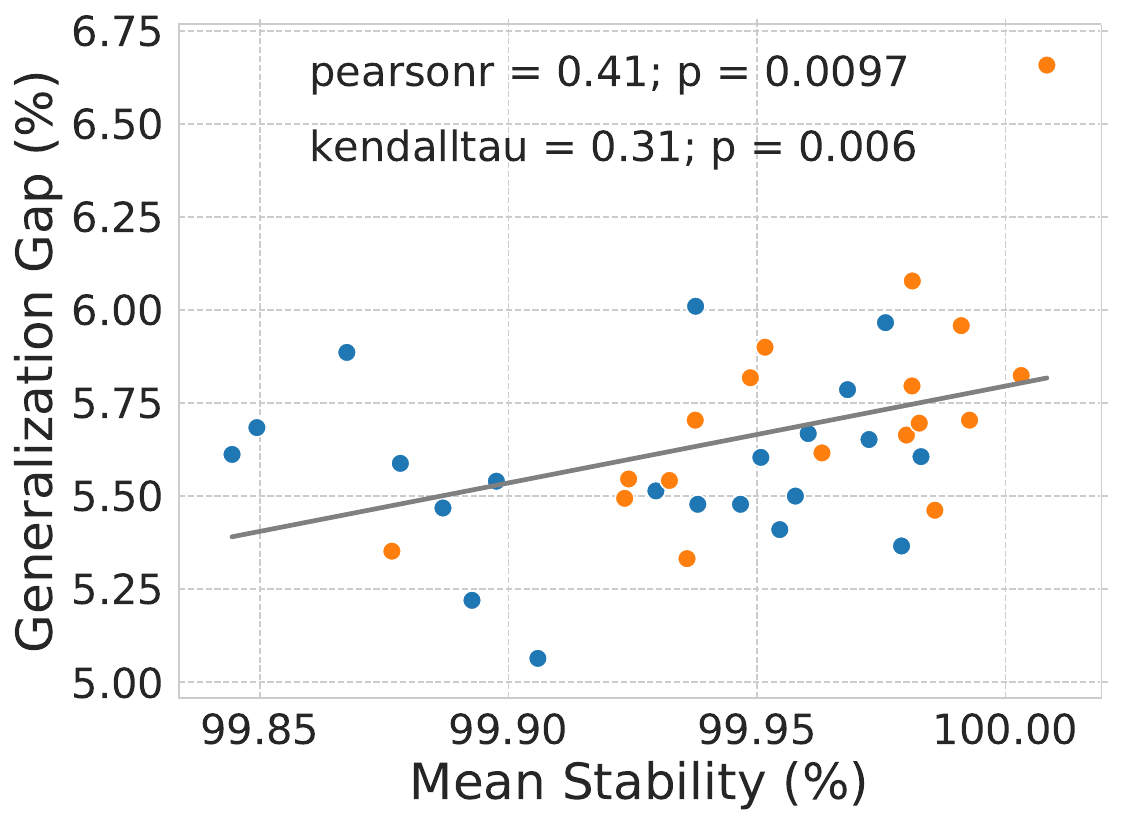}}
    \end{minipage}    
\caption{The generalization stability tradeoff is present among pruned models with high stabilities, which facilitate less disruption of the training procedure, and allows pruning to outperform the baseline model. The blue dots are runs with the less stable (99.93\% average stability) configuration that pruned 10\% of each layer, while the orange dots show runs from the more stable (99.96\% average stability) configuration that pruned 6\% of each layer.}
\label{fig:beat_baseline}
\end{figure}

Consistent with the generalization-stability tradeoff's broad presence and ability to improve on the baseline at higher stabilities, we found that, while the method that only pruned 6\% of each layer led to 94.25\% average test accuracy (less than the baseline), the less stable method that pruned 10\% of each layer led to 94.41\% accuracy, outperforming the baseline (see Figure \ref{fig:beat_baseline} for the generalization level and gap attained on each pruned-model run). While the improvement here is small, we stress that we are using a very simple pruning scheme that prunes a constant fraction of each layer and we did not search for an optimal configuration, suggesting that further improvements are possible.

Importantly, this result also shows that a group of methods that is more unstable on average (e.g., those in Figure \ref{fig:resnetM} compared to those in Figure \ref{fig:beat_baseline}) does not necessarily generalize better, even when the generalization-stability tradeoff is present within each set of methods. Here, this may have been caused by low pruning stability reducing the benefit of the regularization (weight decay) already being applied to the model, which would make it beneficial to prune in a higher stability regime that is less disruptive.

\section{Pruning noise details, visualization, and more results}
\label{sec:app_noise}

\begin{figure}[h]
  \centering
     \centerline{\includegraphics[width=.9\linewidth]{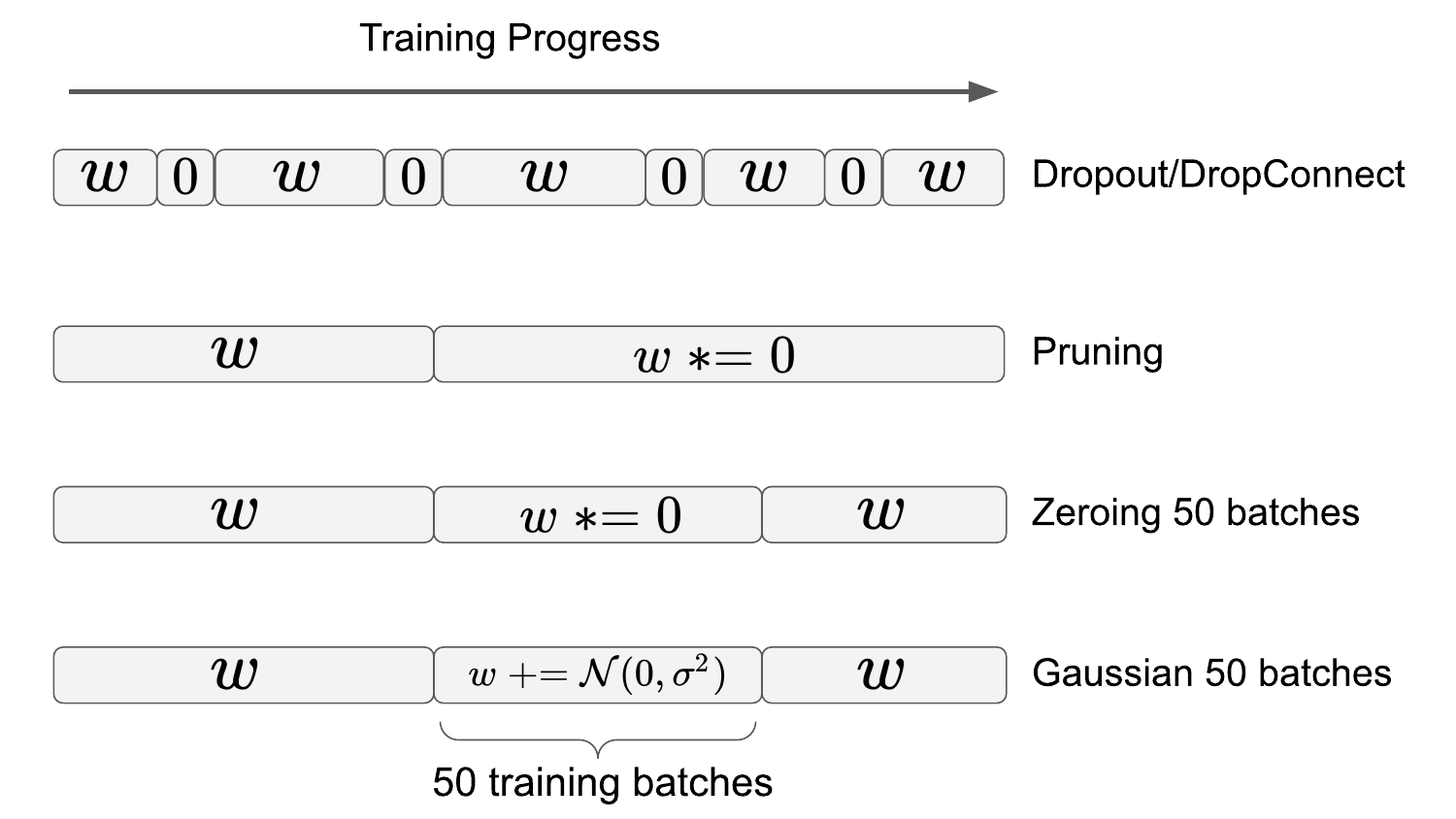}}
\caption{The effect of pruning on a given weight can be likened to that of Dropout/DropConnect \cite{hinton2012improving,srivastava2014dropout,wan2013regularization}, multiplicative zeroing noise (``Zeroing''), and additive Gaussian noise injection (``Gaussian'').}
\label{fig:noise_viz}
\end{figure}

\subsection{Figure \ref{fig:noise} configuration}
\label{sec:details_noise}

In Figure \ref{fig:noise}, pruning/noise-injection targeted the final four convolutional layers of VGG11 during training with (layerwise) starting epochs $s=(3,4,5,6)$, ending epochs $e=(150,150,150,275)$, pruning fractions $p=(0.3,0.3,0.3,0.9)$, and inter-pruning-iteration retrain period $r=40$; we continued using the filter $\ell_2$-norm to score filters. When pruning, we only zeroed the filters, rather than the filters and their associated batch-normalization affine transformation parameters (as done in our other results). When injecting pruning-like noise, we used the same pruning schedule and percentages, but applied noise to the filter weights instead of removing them. Figure \ref{fig:noise_viz} shows an illustration of the similarity between permanently pruning weights $w$ (setting them to zero for the remainder of training), and different kinds of noise injection.

Applying the multiplicative zeroing noise in Figure \ref{fig:noise} entailed multiplying the weights of a filter by zero (before each forward pass) for the specified number of training batches (e.g., once in the case of "Zeroing 1"). The temporary zeroing would still effectively remove any weights that do not learn after reentering the model. However, we observed that pruning all reentered weights at convergence resulted in a marked drop in performance (for all noise schemes except ``Zeroing 1105''), showing that the reentered weights had typically learned after reentry, and that temporary zeroing is therefore less harmful to capacity than permanent pruning. We constructed a variation of this analysis, shown in Section \ref{sec:variation_noise}, that allows weights back in at their pre-zeroing values; this variation also finds that permanent pruning is not necessary and pruning reentered weights always leads to a performance drop.

The Gaussian noise added to weights had mean $0$ and standard deviation equal to the empirical standard deviation of an unmodified filter from the same layer. Our experiments run 391 training batches, 79 test batches, a pruning/noising iteration (even on epochs when no parameters are pruned/noised), then 79 more test batches (to compute stability). We added Gaussian noise to filter parameters on each batch (training or test) until the specified number of training batches was reached.  As such, to add Gaussian noise to the parameters for 50 training batches (`Gaussian 50'' in Figure \ref{fig:noise}), we first added Gaussian noise for the 79 test batches following the pruning/noising iteration. Since the parameters were not updating during this time, this is equivalent to adding Gaussian noise once using a variance 79 times the variance of an unmodified filter, then once for each of the next 50 training batches using noise with a variance equal to the variance of an unmodified filter. ``Gaussian 1'' in Figure \ref{fig:noise} had noise added on just the first test batch (providing an identical effect to adding the noise on the first training batch), and ``Gaussian 1105'' in Figure \ref{fig:noise} had noise added on all training and test batches until 1105 training batches were reached.

The models were trained on CIFAR-10 with Adam for 325 epochs with $lr_s = (150,300)$. The error bars are 95\% confidence intervals for the means, bootstrapped from 10 distinct runs of each experiment.

\subsection{Variation of Figure \ref{fig:noise} with weights reentering at original values}
\label{sec:variation_noise}

\begin{figure}[!t]
    \centering
    \begin{minipage}[!t]{.8\linewidth}
      \centering
  \centerline{\includegraphics[width=\linewidth]{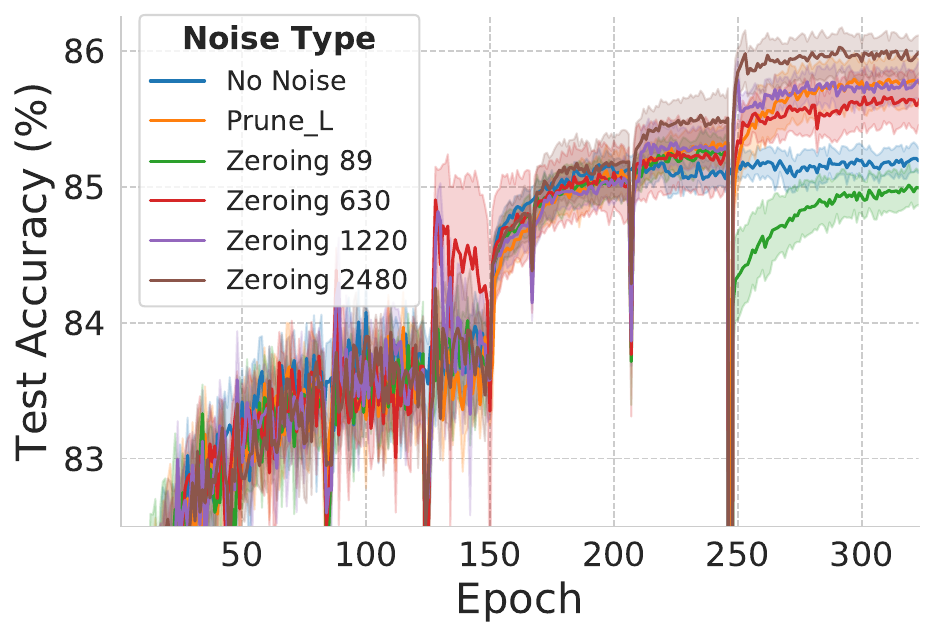}}
    \end{minipage}    
\caption{Generalization improvements from pruning bear resemblance to those obtained by using temporary multiplicative zeroing with the zeroed weights reentering at their pre-zeroing values, as long as the noise is applied for enough batches/steps.}
\label{fig:noise_alt}
\end{figure}

In Figure \ref{fig:noise_alt} we applied the temporary zeroing noise to both filter weights and the corresponding batch normalization affine transformation parameters (in Figure \ref{fig:noise}, the batch normalization parameters were not modified). However, we first stored the values associated with these parameters, as well as the batch normalization running mean and standard deviation. After the prescribed number of batches of zeroing was completed, we restored these variables to their pre-zeroing original values. A further modification we made was keeping track of which filters had been zeroed at any point, and zeroing all filters that had ever been zeroed at each ``pruning'' iteration, which created less stable zeroing events.\footnote{ In future work, it might be interesting to look at the properties of the two subnetworks that are created by this process (the subnetwork that was noised at some point, and the subnetwork that never had zeroing applied to its elements).} Note that the network never stops relying on the weights that were once zeroed (i.e., for ``Zeroing 2480'' in Figure \ref{fig:noise_alt}, accuracy falls at the end of training if all previously-zeroed weights are removed, which wasn't the case for ``Zeroing 1105'' in Figure \ref{fig:noise}, as discussed in Section \ref{sec:details_noise}). 

\section{Flatness}
\label{sec:app_flat}

\subsection{Figure \ref{fig:flatness} configuration and details}

We used the same VGG training and pruning configuration discussed in Section \ref{sec:details_corr}. We added one pruning approach with a random target and $r=40$, and another pruning approach that pruned the model before training began (``Scratch Pruning'') \cite{li2016pruning,liu2017learning}.

We measure \textit{curvature} using an approximation to the Hessian $\mathbf{H}$ trace: we sum the first 100 eigenvalues of the Hessian, though our results are unaffected by whether we use this approach or only sum the eigenvalues larger than the spectral radius times a small factor  \cite{thomas2019interplay}.  We measure \textit{noise} with the uncentered gradient covariance $\mathbf{C}$ trace. While the centered gradient covariance matrix provides information about sensitivity of $\nabla L_w$ to the sample, the uncentered gradient covariance matrix $\mathbf{C}$ that we compute should be similar to its centered counterpart near a minimum \cite{thomas2019interplay}, so we describe the gradient covariance as providing of information about the sensitivity of $\nabla L_w$ to the sample.

\subsection{Computing the Hessian eigenvalues and gradient covariance}
\label{sec:computeH}

\begin{figure}[h]
  \centering
    \begin{minipage}[h]{.9\linewidth}
      \centering
     \centerline{\includegraphics[width=\linewidth]{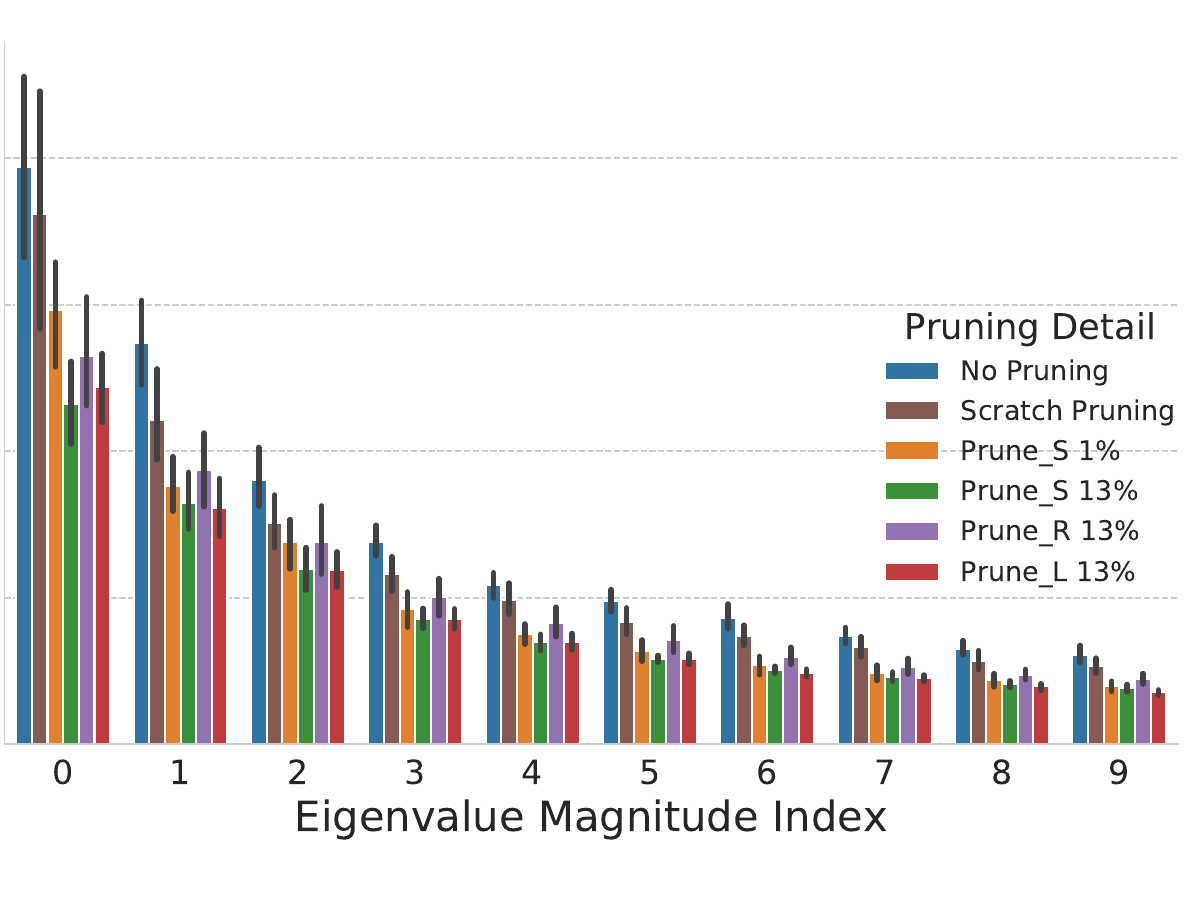}}
    \end{minipage}
\caption{First ten eigenvalues of the Hessian of the test loss.}
\label{fig:hessian_10}
\end{figure}

\begin{figure}[t]
    \begin{minipage}[h]{.49\linewidth}
      \centering
     \centerline{\includegraphics[width=\linewidth]{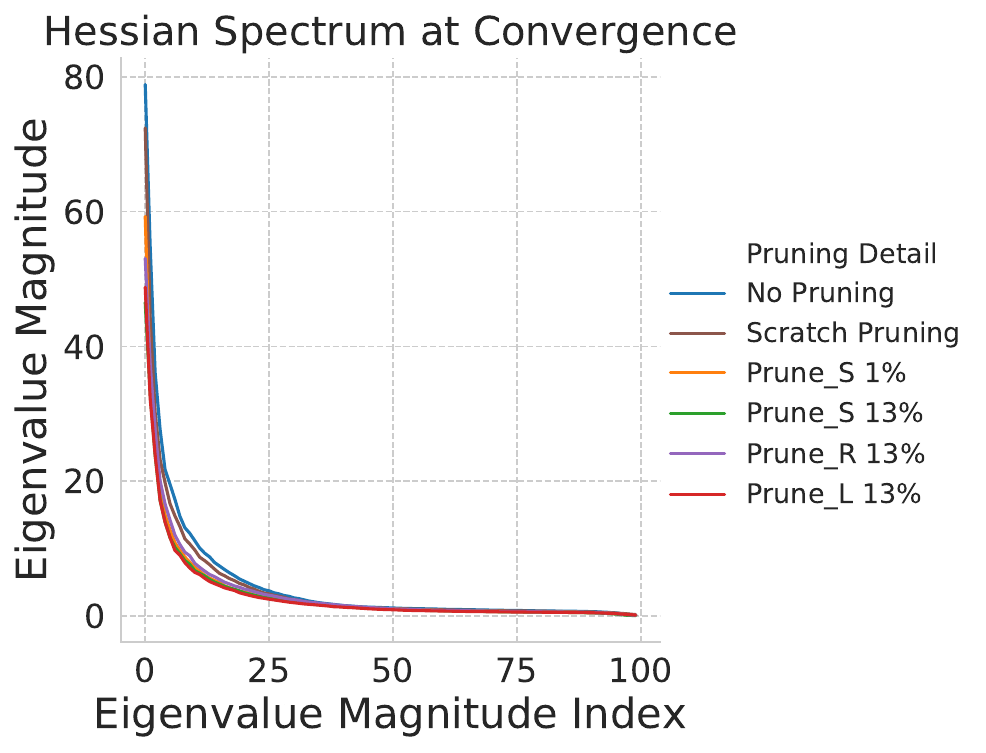}}
    \end{minipage}
    \begin{minipage}[h]{.49\linewidth}
      \centering
     \centerline{\includegraphics[width=\linewidth]{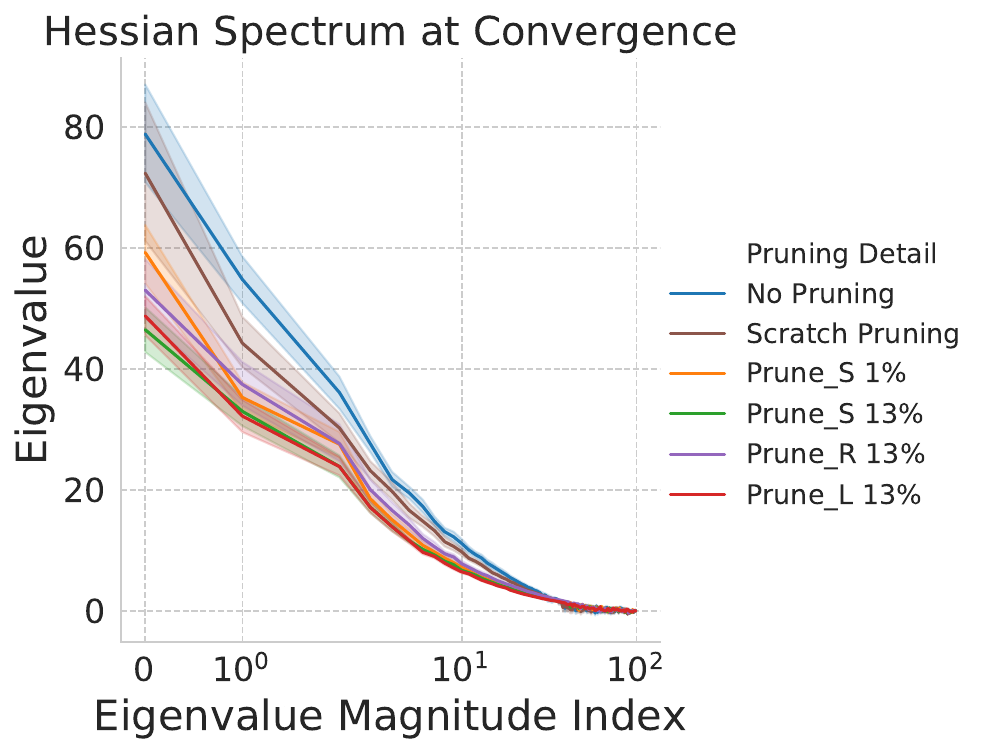}}
    \end{minipage}
\caption{The first 100 eigenvalues of the Hessian of the test loss.}
\label{fig:hessian_spectra}
\end{figure}

We estimate $\mathbf{H}$ and $\mathbf{C}$ using a subset of 512 \textit{test} data samples \cite{thomas2019interplay} on epoch 315. To compute the first 100 eigenvalues of the Hessian, we use the power method with deflation provided by \cite{hessian-eigenthings}, iterating 100 times or until a tolerance of $0.0001$ is reached for each eigenvalue. The first ten eigenvalues (each with a 95\% confidence interval for the mean based on bootstrapping with the 10 runs per configuration) are shown in Figure \ref{fig:hessian_10}. All 100 eigenvalues are shown in Figure \ref{fig:hessian_spectra}.

\subsection{Other flatness measures and results}

\begin{figure}[h]
  \centering
    \begin{minipage}[h]{.45\linewidth}
      \centering
     \centerline{\includegraphics[width=\linewidth]{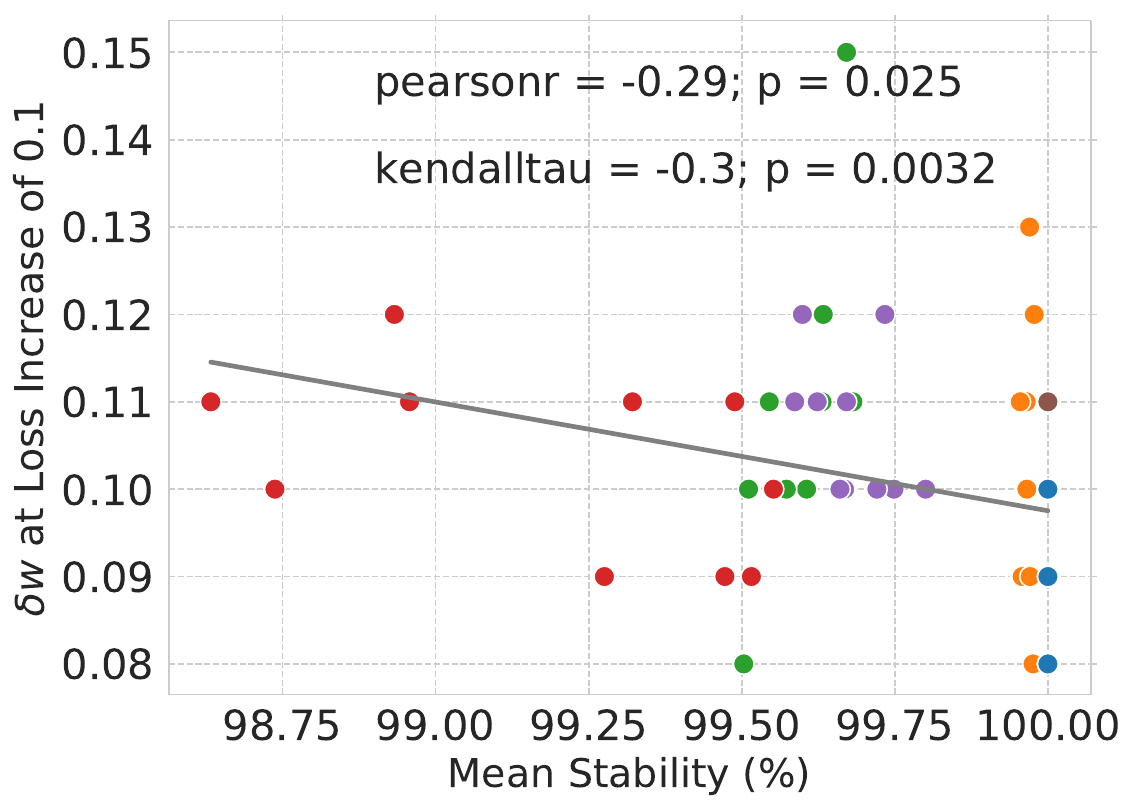}}
    \end{minipage}
    \begin{minipage}[h]{.45\linewidth}
      \centering
     \centerline{\includegraphics[width=\linewidth]{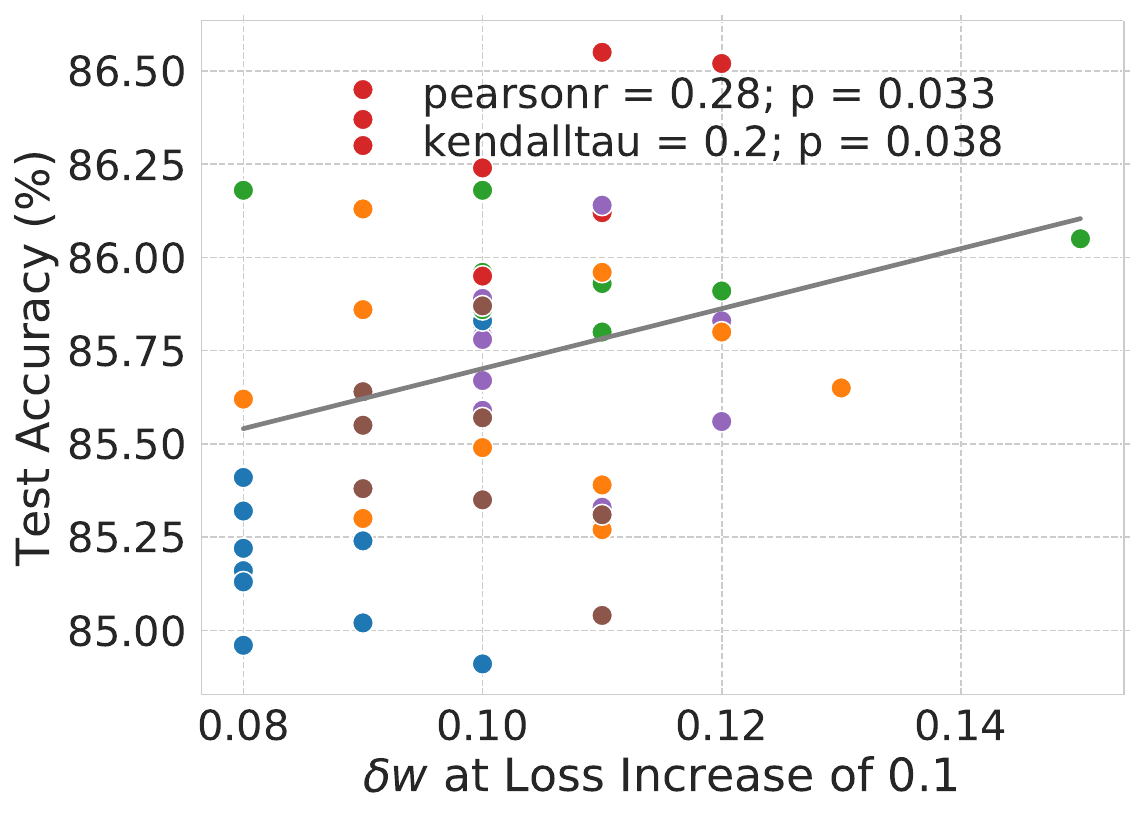}}
    \end{minipage}
  \centering
    \begin{minipage}[h]{.99\linewidth}
      \centering
        \centerline{\includegraphics[width=\linewidth]{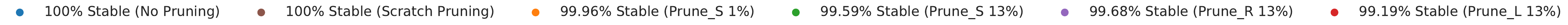}}
    \end{minipage}
\caption{When perturbing the parameters in the neighborhood of an optimum, we find that the methods trained with less stability can sustain larger perturbations to the weights before reaching a loss increase of 0.1.}
\label{fig:epsilon}
\end{figure}

\begin{figure}[!h]
  \centering
    \begin{minipage}[h]{.9\linewidth}
      \centering
     \centerline{\includegraphics[width=\linewidth]{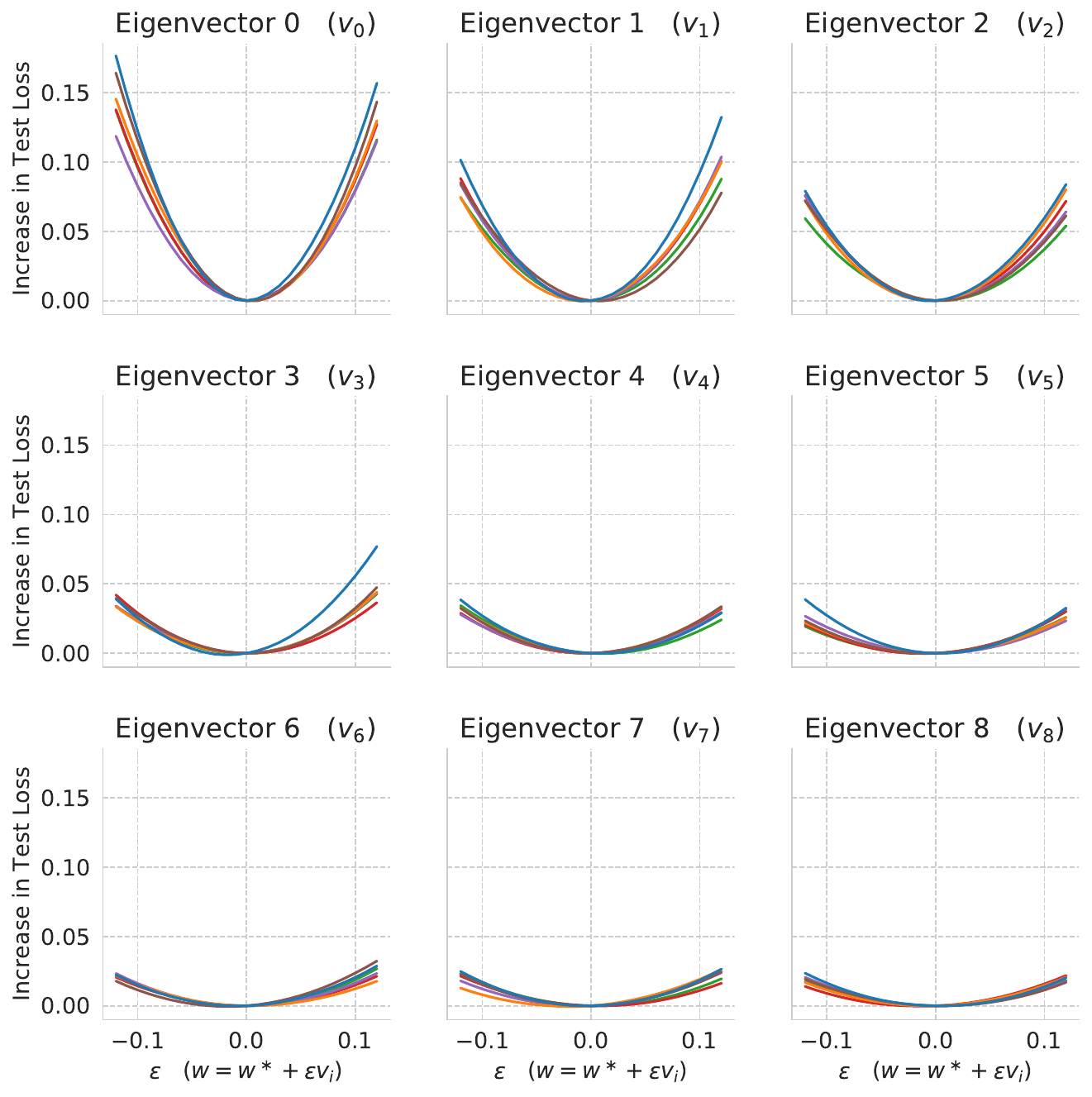}}
    \end{minipage}
  \centering
    \begin{minipage}[h]{.99\linewidth}
      \centering
        \centerline{\includegraphics[width=\linewidth]{figures_and_tables/PDFs/Figure_5_Legend.pdf}}
    \end{minipage}
\caption{Parameter perturbations in the directions of more dominant Hessian eigenvectors cause greater increases in the test loss for a given $\varepsilon$.}
\label{fig:perturb}
\end{figure}

Inspired by \cite{yao2018hessian,jiang2019fantastic}, we used the Hessian eigenvectors $v_i$ that we had computed via the power method (as described in Section \ref{sec:computeH}) to perturb the parameters and measure how the loss changes in a neighborhood of the minimum $w^\ast$. Specifically, we compute the test loss at the point $w=w^\ast + \varepsilon v_i$ (Figure \ref{fig:perturb}). The distance $\varepsilon$ that could be used before a loss increase of $0.1$ was reached (for any $v_i$) is shown as a function of stability and as a predictor of generalization in Figures \ref{fig:epsilon} and \ref{fig:perturb}.  One drawback of our approach is that we incremented the value of $\varepsilon$ by $0.01$, leading to a less precise estimate of the particular value at which the loss increases by $0.1$.

Perturbing the parameters in the direction of the Hessians' eigenvectors \cite{yao2018hessian} is a kind of worst-case perturbation, to the extent that the weights are at a minimum of the test loss and the gradient is zero. Several, more sophisticated approaches to this perturbation analysis were used in \cite{jiang2019fantastic} and could be useful here. For example, it would be interesting to extend our analysis to include flatness measures derived from PAC Bayesian generalization bounds \cite{jiang2019fantastic}.

\vspace{.5cm} 

\vspace{1.5cm} 

\clearpage

\begin{figure}[t]
  \centering
    \begin{minipage}[h]{.45\linewidth}
      \centering
     \centerline{\includegraphics[width=\linewidth]{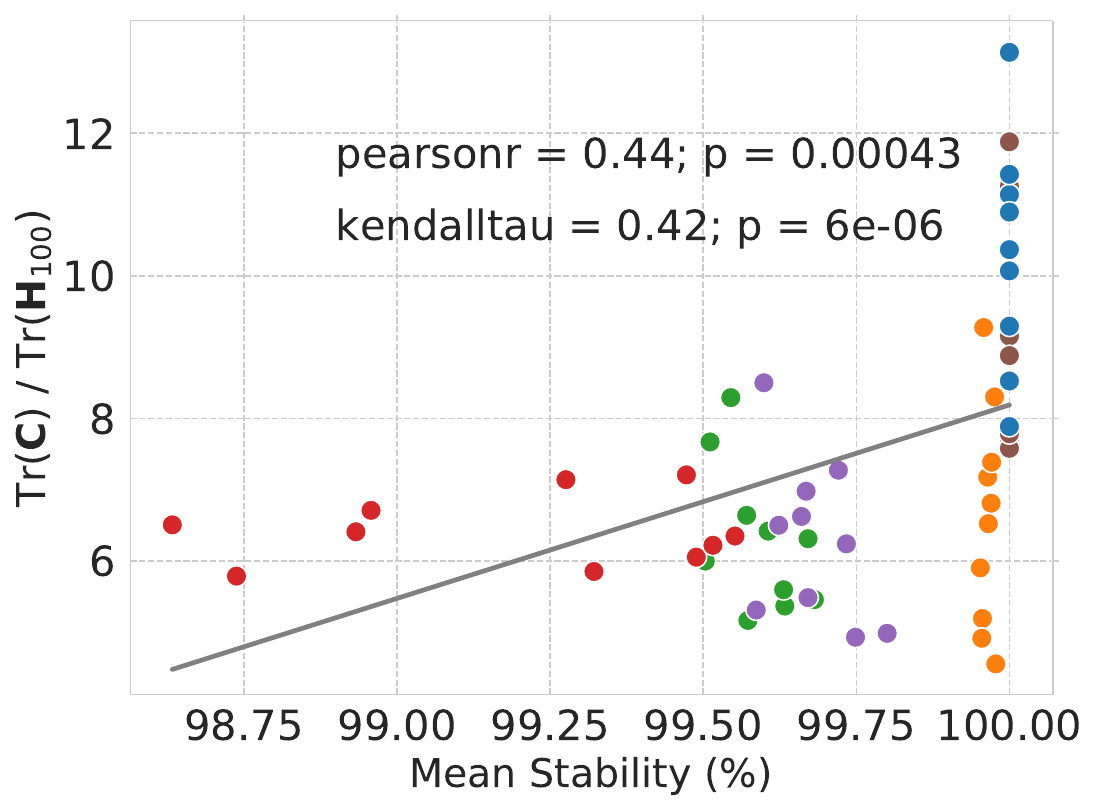}}
    \end{minipage}
    \begin{minipage}[h]{.45\linewidth}
      \centering
     \centerline{\includegraphics[width=\linewidth]{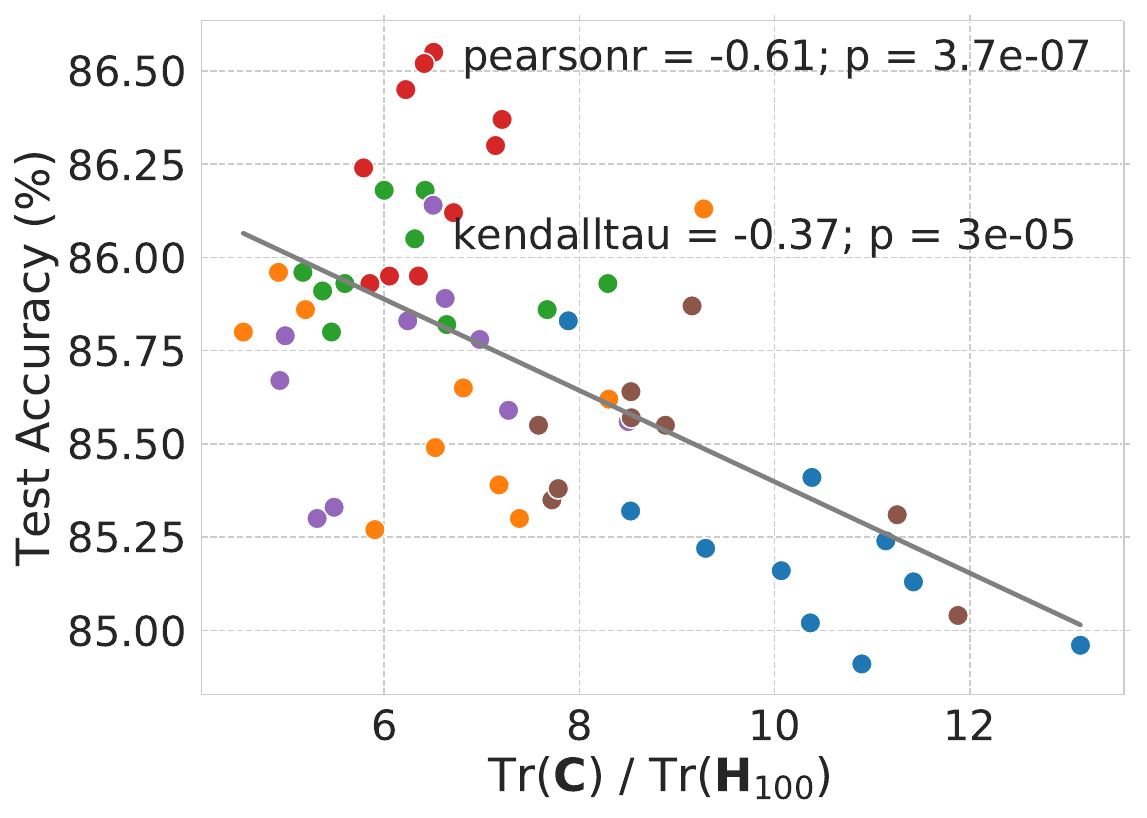}}
    \end{minipage}
  \centering
    \begin{minipage}[h]{.99\linewidth}
      \centering
        \centerline{\includegraphics[width=\linewidth]{paper/figures/flatness_legend.pdf}}
    \end{minipage}
\caption{The proxy to the TIC suggests the model will generalize worse as pruning stability rises (left), and it is predictive of generalization (right).}
\label{fig:tic}
\end{figure}

Lastly, we examine our results with respect to a proxy for the Takeuchi Information Criterion \cite{takeuchi1976distribution}, an estimator of the generalization gap \cite{thomas2019interplay} built from $\mathbf{C}$ and $\mathbf{H}$. The proxy we use is a modification of the suggested proxy $\mathrm{Tr}(\mathbf{C})/\mathrm{Tr}(\mathbf{F})$ in \cite{thomas2019interplay}. Specifically, we use $\mathrm{Tr}(\mathbf{C})/\mathrm{Tr}(\mathbf{H})$ instead, which is based on the same reasoning given for $\mathrm{Tr}(\mathbf{C})/\mathrm{Tr}(\mathbf{F})$ in \cite{thomas2019interplay}. Here, too, we approximate $\mathrm{Tr}(\mathbf{H}$) using the first 100 eigenvalues. This TIC proxy accurately describes generalization levels (Figure \ref{fig:tic} right) and suggests models will generalize better as stability decreases (Figure \ref{fig:tic} left).

\begin{figure}[h]
  \centering
    \begin{minipage}[!t]{.8\linewidth}
      \centering
        \centerline{\includegraphics[width=\linewidth]{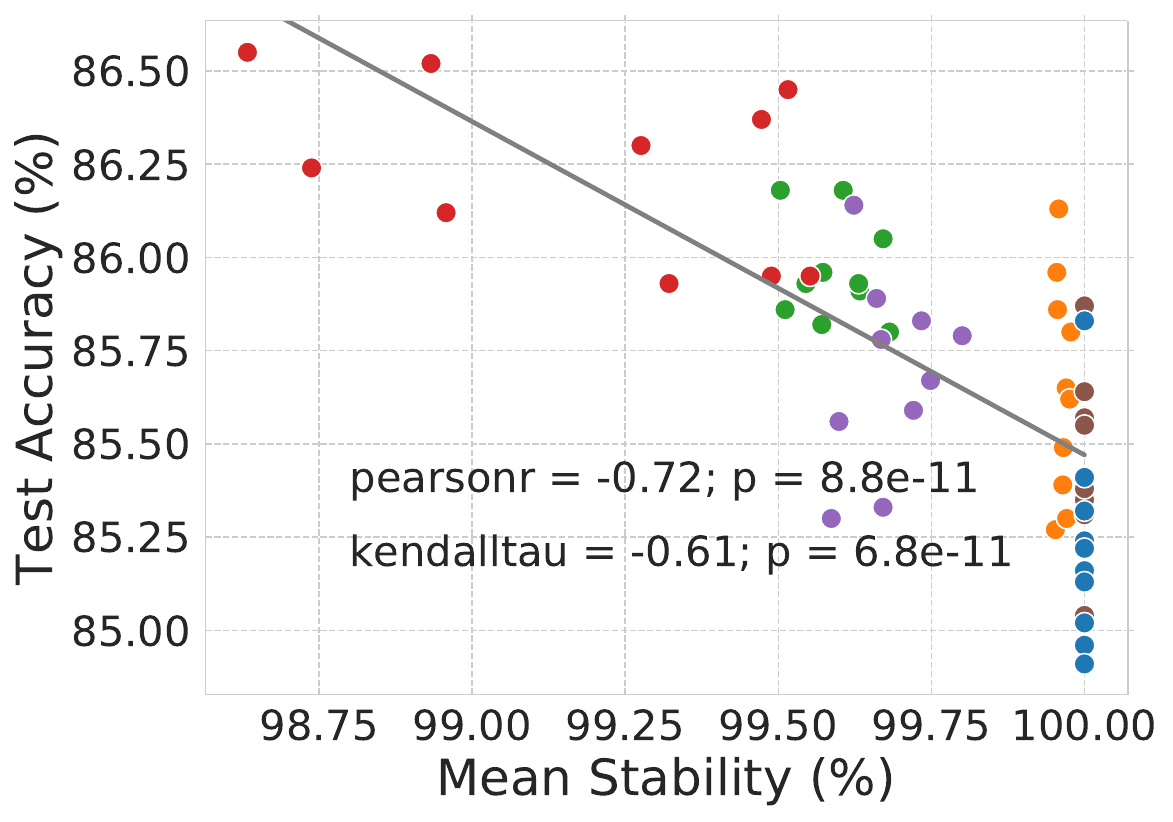}}
    \end{minipage}
  \centering
    \begin{minipage}[h]{.99\linewidth}
      \centering
        \centerline{\includegraphics[width=\linewidth]{paper/figures/flatness_legend.pdf}}
    \end{minipage}
\caption{Lower stability is associated with higher generalization in the models we analyzed the flatness of.}
\label{fig:stability_acc}
\end{figure}

The accuracy and stability of the various approaches we analyzed the flatness of are displayed in Figure \ref{fig:stability_acc}.

\section{The generalization-stability tradeoff with CIFAR100}
\label{sec:app_c100}

In our experiments we saw the presence of a generalization-stability tradeoff in networks trained on CIFAR10. However, it's unclear whether this phenomenon will be present when we move to larger datasets. One possibility is that the tradeoff was an artifact of pruning models trained on CIFAR10, rather than a more general phenomenon. Alternatively, the tradeoff may exist when pruning models trained on various datasets.

To test this, we trained ResNet18 on CIFAR100 using no pruning, scratch pruning, stable pruning during training, and unstable pruning during training. If the tradeoff was an artifact of our use of CIFAR10, then we would not expect to see a generalization-stability tradeoff in these results. On the other hand, if the tradeoff is a more general phenomenon, then we should see it in this experiment. 

Consistent with the tradeoff applying to data other than CIFAR10, we found that reduced pruning stability helps generalization of ResNet18 trained on CIFAR100 (Table \ref{table:c100}). Despite only using three runs per configuration, the improvements of the less stable Prune$_\mathrm{L}$ method over both Scratch Pruning and Prune$_\mathrm{S}$ are statistically significant at less than the 5\% significance level (using a two-tailed t-test).
\\
\begin{table}[h]
\caption{Benefit of low stability for CIFAR100 at 15\% total pruning percentage}
\label{table:c100}
    \centering
    \begin{tabular}{ccccc}
\\
        \toprule
        \multirow{1}{*}[-1.5em]{Method} & 
         \multicolumn{2}{c}{Test Accuracy} & \multicolumn{2}{c}{Stability}\\
        \cmidrule{2-3} \cmidrule{4-5} \\
         {}   & Mean (\%) & Std. Dev. & Mean (\%) & Std. Dev. \\
        \midrule
        No Pruning & 73.28  & 0.12 & 100  & N/A \\
        Scratch Pruning & 73.11  & 0.13 & 100  & N/A
         \\
        Prune$_\mathrm{S}$ &  73.22  & 0.09 & 91.94  & 4.10 \\
        Prune$_\mathrm{L}$ & \textbf{73.41}  & 0.08 & 86.98  & 6.87 \\
        \bottomrule
    \end{tabular}
\\
\end{table}

We used an experimental design inspired by \cite{nakkiran2019deep}, in which the models display worsening generalization as ResNet18's width parameter is reduced. Specifically, we used 4000 training epochs, data augmentation, and an initial learning rate of 0.0001 with the Adam optimizer. Unlike \cite{nakkiran2019deep}, we reduced the learning rate to one-tenth its initial value at epoch 2000 ($lr_s = (2000)$), which raised the generalization level of all methods examined. 

We pruned a total of $15\%$ of the model by pruning the final convolutional layer twice during training with starting epoch $s=(2500)$, ending epoch $e=(3250)$, inter-pruning retrain period $r=750$, and pruning fraction $p=(0.7)$. We used the same scoring method that we used for ResNet18 in Section \ref{sec:tradeoff_bn}, though we obtained similar results when scoring using the empirically-calculated average of the absolute value of the post-non-linearity activations (Prune$_\mathrm{L}$ had the same mean test accuracy, and Prune$_\mathrm{S}$ mean test accuracy went down .03\% to 73.19\%). We initially tried pruning more of the network, but doing so quickly reduced generalization relative to the baseline (as shown in \cite{nakkiran2019deep}), indicating the difficulty of obtaining generalization benefits by pruning models trained on larger datasets.

\end{document}